\documentclass[acmlarge]{acmart}
\usepackage{tikz}
\usepackage{amsmath}

\usepackage[ruled,linesnumbered,vlined]{algorithm2e}
\usepackage{subfigure}
\usepackage{array}
\usepackage{enumitem}
\usepackage{graphicx}

\usepackage{footnote}
\usepackage{makecell}
\usepackage{multicol}
\usepackage{multirow}
\usepackage{booktabs}
\usepackage{tabularx}
\usepackage{hhline}
\usepackage{setspace}
\usepackage{colortbl}
\usepackage[most]{tcolorbox}

\setcopyright{none}
\renewcommand\footnotetextcopyrightpermission[1]{} % removes footnote with conference information in first column
\AtBeginDocument{%
  \providecommand\BibTeX{{%
    \normalfont B\kern-0.5em{\scshape i\kern-0.25em b}\kern-0.8em\TeX}}}

\acmVolume{0}
\acmNumber{0}
\acmArticle{0}
\acmMonth{8}

\begin{document}

\title{ECLM: Efficient Edge-Cloud Collaborative Learning with Continuous Environment Adaptation}

% \affiliation{%

\author{Yan Zhuang}
\affiliation{%
  \institution{Shanghai Jiao Tong University}
  \city{Shanghai}
  \country{China}
}
\email{zhuang00@sjtu.edu.cn}

\author{Zhenzhe Zheng}
\affiliation{%
  \institution{Shanghai Jiao Tong University}
  \city{Shanghai}
  \country{China}
}
\email{zhengzhenzhe@sjtu.edu.cn}

\author{Yunfeng Shao}
\affiliation{%
  \institution{Huawei Noah’s Ark Lab}
  \city{Beijing}
  \country{China}
}
\email{shaoyunfeng@huawei.com}

\author{Bingshuai Li}
\affiliation{%
  \institution{Huawei Noah’s Ark Lab}
  \city{Beijing}
  \country{China}
}
\email{libingshuai@huawei.com}

\author{Fan Wu}
\affiliation{%
  \institution{Shanghai Jiao Tong University}
  \city{Shanghai}
  \country{China}
}
\email{fwu@cs.sjtu.edu.cn}

\author{Guihai Chen}
\affiliation{%
  \institution{Shanghai Jiao Tong University}
  \city{Shanghai}
  \country{China}
}
\email{gchen@sjtu.edu.cn}


\begin{abstract}
To bring the great power of modern DNNs into extensive mobile applications, current practices primarily employ one of the two learning paradigms: cloud-based learning or on-device learning. 
The former leverages abundant computational resources on the cloud to provide high-performance services with large models, while the latter executes small models close to users, enabling fast-response and low-cost model services. 
Despite their own advantages, neither of these two paradigms could effectively deal with highly dynamic edge environments reflected in frequent data distribution shifts and on-device resource fluctuations.
In this paper, we propose ECLM, an edge-cloud collaborative learning framework for rapid model adaptation for dynamic edge environments.
We first propose a novel block-level model decomposition design to decompose the original large cloud model into multiple combinable modules. By flexibly combining a subset of the modules, this design enables the derivation of compact, task-specific sub-models for heterogeneous edge devices from the large cloud model, and the seamless integration of new knowledge learned on these devices into the cloud model periodically. As such, ECLM ensures that the cloud model always provides up-to-date sub-models for edge devices.
We further propose an end-to-end learning framework that incorporates the modular model design into an efficient model adaptation pipeline, including an offline on-cloud model prototyping and training stage, and an online edge-cloud collaborative adaptation stage. 
Extensive experiments over various datasets demonstrate that ECLM improves model performance (\emph{e.g.}, 18.89\% accuracy increase) and resource efficiency (\emph{e.g.}, 7.12$\times$ communication cost reduction) in adapting models to dynamic edge environments by efficiently collaborating the edge and the cloud models.
% short version
% Pervasive mobile AI applications primarily employ one of the two learning paradigms: cloud-based learning (with powerful large models) or on-device learning (with lightweight small models). Despite their distinct advances, neither paradigm effectively handles dynamic edge environments with frequent data distribution shifts and on-device resource fluctuations. In this paper we propose ECLM, an edge-cloud collaborative learning framework for rapid model adaptation for dynamic edge environments. We first propose a novel block-level model decomposition design to decompose the original large cloud model into multiple combinable modules. This design enables the creation of compact, task-specific sub-models for edge devices and the seamless integration of new knowledge from these devices into the cloud model. This ensures that the cloud model always provides up-to-date sub-models to edge devices. We further propose an end-to-end learning framework that incorporates the modular model design into an efficient model adaptation pipeline including an offline on-cloud model prototyping and training stage, and an online edge-cloud collaborative adaptation stage. Extensive experiments over various datasets demonstrate that ECLM improves model performance (\emph{e.g.}, 18.89\% accuracy increase) and resource efficiency (\emph{e.g.}, 7.12$\times$ communication cost reduction) in adapting models to dynamic edge environments by efficiently collaborating the edge and the cloud models.
\end{abstract}

\maketitle

\section{Introduction}

Deep Neural Networks (DNNs) have shown remarkable performance in various applications such as computer vision~\cite{liu2019edge,zeng2017mobiledeeppill}, natural language understanding~\cite{ravi2018self,zhang2018deep},  human activity recognition~\cite{Ouyang2021clusterfl,tu2021feddl}, and etc. Despite their strong capabilities, it usually requires a huge volume of training data, which is usually produced and stored on edge devices (\emph{e.g.}, mobile phones). In the traditional cloud-based learning paradigm, the devices need to upload their data to the cloud with strong computational power to do centralized model training and inference, while this paradigm may suffer from heavy communication overhead~\cite{mcmahan2017communication}, high processing latency~\cite{chengfei2022walle} and privacy leakage problems~\cite{niu2020billion}. 
To fix these issues, an on-device learning paradigm emerges to leverage the growing system capabilities and data volumes on edge devices by enabling model inference and training locally on edge devices, and acts as a complementary to the classic cloud-based learning. 

Although the above learning paradigms have their own advances and applicable scenarios, problems arise when faced with highly dynamic edge environments~\cite{fang2018nestdnn,wang2019adaptive,liu2021adaspring}. Such dynamics are reflected in two aspects. One is that the application context in edge environments could frequently change, leading to shifting local data distributions and varying performance requirements (\emph{e.g.}, different accuracy-latency tradeoffs). %on edge devices.
The other one is that the on-device resources for model execution could also vary dramatically across devices and times. These dynamics require the learning system to fast adapt models 
%on-service continuously 
to maintain a satisfying performance. 

Unfortunately, neither the cloud-based learning paradigm nor the on-device learning paradigm could effectively deal with highly dynamic edge environments, and inevitably suffer from model performance drops~\cite{fang2018nestdnn,wen2023adaptivenet}.
For the cloud-based learning paradigm under dynamic edge environments, the edge devices request the cloud for new local models when changes of environments are detected. However, the cloud model is trained using historical (proxy) data prior to deployment, which can not provide up-to-date models for edge devices, resulting in 11\% accuracy drops demonstrated in our experiments (Section \ref{s2}). Besides, this paradigm would induce prohibitive computation and communication costs to serve huge amount of edge devices and deal with frequently changing edge environments. 
On the other side, for the on-device learning paradigm, edge devices could update their models locally using newly-collected data to adapt to the dynamic edge environments. Nevertheless, they still suffer from sub-optimal model performance with up to 10\% accuracy drops\footnote{when comparing with the ideal situation where an edge model can be updated by collecting enough data.} due to the sparse and biased training data possessed by an individual edge device. Furthermore, the on-device resource competition among model training and inference processes~\cite{bhardwaj2022ekya,fang2018nestdnn,Padmanabhan2022gemel} could lead to $5.06\times$ prolonged model response latency. 

To tackle the drawbacks of merely cloud-based or edge-based learning paradigm in dynamic edge environments, in this paper, we propose ECLM, an edge-cloud collaborative learning framework
%~\cite{chengfei2022walle,yao2022edge,bonawitz2019towards,yao2021device,Kang2017Neurosurgeon} 
to support rapidly adapting models for dynamic edge environments. 
Within this paradigm, ECLM takes advantage of both cloud (\emph{e.g.}, plenty of resources) that maintains a large and powerful model for superior integrated performance, and edge (\emph{e.g.}, close to users and data sources) that employs compact and specialized sub-models for agile executing and updating. The cloud is responsible for aggregating and storing the continuous new learned knowledge, and edge devices retrieve sub-models from the cloud to execute and also collect new knowledge from encountering  environments.  
Based on this intuitive idea, ECLM aims to provide high-performance and fast-adaptation learning simultaneously to deal with frequently changing edge environments. 
The key to collaborating between edge and cloud relies on two critical steps: one is to efficiently derive personalized sub-models from the large cloud model for executing on resource-limited edge devices, and the other one is to integrate new knowledge learned on heterogeneous edge devices back into the cloud model. The cloud-to-edge model derivation enables lightweight edge models to obtain specialized abilities from the cloud on demand for encountering edge environments, and the edge-to-cloud model aggregation further enhances the ability of the large cloud model to deal with the new edge environments.

To complete these two steps, two challenges arise in {deriving personalized edge models} from the large cloud model for on-device training and inference, and {aggregating heterogeneous edge models} for effective knowledge transfer from edge devices to the cloud. 
(i) The first challenge comes from that edge devices 
have 
limited hardware resources and non-IID data distributions~\cite{fang2018nestdnn,farcas2022model,diao2021heterofl}. 
Due to the limitation of system resources, 
edge devices could not afford to train a full large model. 
The non-IID data distribution can reflect that 
the local task of an edge device (\emph{e.g.}, recognizing a subset of target objects) is a sub-task of the global task (\emph{e.g.}, recognizing all potential target objects), which calls for a personalized sub-model instead of a general global model. 
Therefore, the sub-models need to have compact sizes (for limited on-device resources) and specialized abilities (to deal with target local tasks), which are non-trivial to achieve simultaneously. The existing works~\cite{horvath2021fjord,alam2022fedrolex,hong2022efficient,Bouacida2021adaptive}, extracted sub-models using strategies such as ordered dropout~\cite{horvath2021fjord} or rolling sub-model extraction~\cite{alam2022fedrolex}, which only considers the limited on-device resources and ignores the specialized abilities of sub-models. 
(ii) The second challenge 
comes from that the edge models are heterogeneous in both structures and parameters, introducing difficulties in integrating knowledge from edge models to the cloud. On the one hand, although knowledge distillation~\cite{hinton2015distilling,he2020group} allows transferring knowledge between models with heterogeneous structures, it imposes extra storage and computation burden on edge devices, and is also time-consuming due to its re-training process.
On the other hand, simply averaging parameters (\emph{e.g.}, FedAvg~\cite{mcmahan2017communication}) could lead to negative impacts on effective knowledge transfer due to parameter conflicts~\cite{modeling2018ma,crossstitch2016misra}, as the edge models are trained on diverse local tasks independently. 
%(iii) The final challenge is that 
In addition, frequently changing and complex edge environments further exacerbate these two challenges by 
raising high efficiency requirements to quickly adapt to encountering new edge environments. The existing works that allow edge devices to adapt local models for flexible accuracy-latency
tradeoff on-demand (\emph{i.e.}, the ``one-shot training and deployment'' design)~\cite{fang2018nestdnn,han2021legodnn,wen2023adaptivenet}
are hard to satisfy the above requirements as their model parameters are essentially not updated during model serving. 

The core idea of ECLM to solve the above challenges is a novel modular model decomposition design, based on which we can efficiently derive personalized sub-models for edge devices, and effectively aggregate the updated sub-models to integrate new-learned knowledge of edge models into the cloud model, and thus the cloud model is able to provide up-to-date sub-models for edge devices in return. Specifically, ECLM decomposes the large cloud model into multiple well-separated but combinable modules. In its essence, ECLM 
%not only decomposes the large cloud model in a structural and modular way, but 
intrinsically decomposes the global task (represented by the global data distribution) to multiple sub-tasks (represented by the local data distributions on edge devices), each of which can be solved by a sub-model built by intentionally combining a proper subset of the modules.
With this modular design, we can flexibly derive and aggregate personalized sub-models with diverse model sizes and specialized abilities.

Based on the above idea, we further propose an end-to-end learning framework that incorporates the modular model design into an efficient model adaptation pipeline against dynamic edge environments. This learning framework comprises an offline on-cloud model training stage and an online edge-cloud collaborative adaptation stage. 
In the offline stage, we modularize the cloud model and design a unified module selector
%, which are first trained on the cloud 
to learn model/task decomposition strategies (\emph{i.e.}, how to decompose the global model/task to modules/sub-tasks) and to associate specific sub-tasks to modules.
In the online stage, ECLM efficiently derives personalized sub-models from the cloud model regarding edge devices' local tasks/data distributions\footnote{In this work, we interchangeably use the term local task/local distribution and edge model/sub-model.} and available resources. During serving on edge devices, the sub-models are periodically updated using fresh data, and are further aggregated into the cloud model in a module-wise manner with minimal parameter conflicts. By efficiently collaborating between the edge and the cloud, we are able to organize edge devices and the cloud to continuously update their models to adapt to the dynamic edge environments.

We summarize the contributions in this work as follows: 

\begin{itemize}
\item We propose a novel modular model design to decompose the large cloud model into multiple well-separated but combinable modules, based on which various personalized edge models can be derived on demand and can also be aggregated with minimal parameter conflicts.  

\item We design ECLM, an edge-cloud collaborative learning framework, for continuous model adaptation against dynamic edge environments.
From cloud to edge, we efficiently derive new personalized edge models from the cloud model, respecting the encountering new local data distribution and available on-device resources. From edge to cloud, we aggregate updated edge models to form a new cloud model. %used in the future. 

\item We implemented ECLM on a simulation platform and a real-world testbed with 20 heterogeneous edge devices, and evaluated ECLM over three representative applications: mobile sensing, image classification, and speech recognition with four different datasets and models.
The evaluation results demonstrate the superiority of ECLM in adapting to the dynamic edge environments, 
%with on-device resource fluctuations and local data distribution shifts
achieving up to 18.89\% accuracy improvement and 7.12 $\times$ communication cost reduction.
\end{itemize}

The rest of the paper is organized as follows. Section \ref{s2} introduces the background on edge-cloud collaborative learning, and motivates our work followed by the analysis of design challenges. Section \ref{overview} overviews the ECLM system and general workflow. Section \ref{s4} and Section \ref{s5} elaborates the offline on-cloud model prototyping and training (\emph{i.e.}, model modularization and training methodology) and online edge-cloud collaborative adaptation (\emph{i.e.}, personalized edge model derivation and updated edge model aggregation), respectively. Section \ref{s6} presents the complete framework with our system implementations, and the experimental results are detailed in Section \ref{s7}. The related works are presented in Section \ref{related work}. Finally, Section \ref{s9} concludes this work. 

\section{Background and Motivation \label{s2}} 

\subsection{Edge-cloud Collaborative Learning}

The edge-cloud collaborative learning paradigm is introduced to overcome the limitations of traditional cloud-based learning in terms of communication overhead, high latency and privacy concerns~\cite{yao2022edge,yao2021device,chengfei2022walle}. 
A representative example within this paradigm is federated learning (FL)~\cite{li2020federated,mcmahan2017communication,liu2022distfl,li2023hierachical}, which jointly trains a global model with massive edge devices and only communicates model parameter updates between cloud and edge devices for collaboration, keeping private data on-device. Another example is split learning (SL)~\cite{Gupta2018DistributedLO,Vepakomma2018SplitLF}, the cloud and edge devices train different parts of a complete model, and exchange the features or gradients of latent representations for collaboration.
This paradigm allows edge devices and the cloud to work collaboratively to accomplish learning tasks by exploiting the advances of both edge devices and the cloud simultaneously. For edge device side, considering that more and more data is produced and stored on edge devices, the model inference and training are deployed on-device for desirable model personalization and low-latency response. For cloud side, the cloud having sufficient resources undertakes resource-intensive tasks such as model prototyping and pre-training, and is also responsible for coordinating edge devices under a global view to promote the efficiency of the collaborative learning. 

However, current practices lack efficient collaboration between the cloud and edge models, and 
mainly follow a ``one-shot training and deployment once for all'' paradigm. 
Such static models 
%for deployment 
could not deal with dynamic edge environments~\cite{wang2021context,liu2021adaspring}.
To be specific, edge environments are changing dynamically and frequently, reflected in two aspects: (i) outer environment dynamic: this causes the changes of application context (\emph{e.g.}, varying lighting conditions of a camera or varying usage patterns of edge devices over time), leading to shifting data distributions and varying model performance requirements. 
(ii) inner runtime environment dynamic: there might have multiple applications co-running on an edge device competing for available resources, which leads to resource fluctuation and then unstable local processing time and communication latency. Ignoring these kinds of dynamics will lead to the degradation of system performance. 

\begin{figure}
    \centering    
    \includegraphics[width=0.7\linewidth]{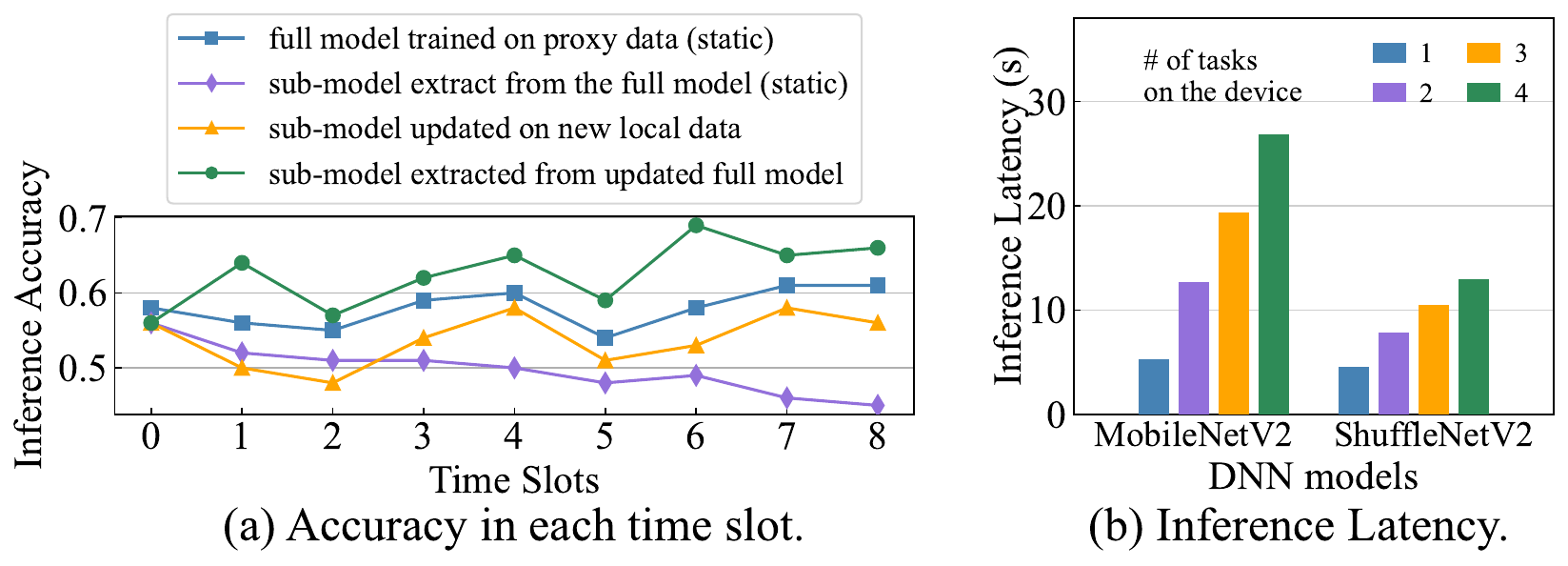}
    % \vspace{-0.8cm}
    \caption{Impact of dynamic edge environments in terms of model accuracy and inference latency using CIFAR100 dataset and VGG16 model. The experiments are conducted on the NVIDIA Jetson Nano device. The results show that either purely cloud-based or on-device learning paradigm suffers from performance degradation in the context of changing environments.}
    \label{impact_dynamic_env}
\end{figure}

We conduct experiments to illustrate the impact of dynamic edge environments. 
For outer environment dynamic, 
Figure~\ref{impact_dynamic_env}(a) shows on-device model accuracy with different adaptation approaches.
%to deal with dynamic edge environments. 
The original dataset is split into a proxy dataset on the cloud to train the cloud model, and an edge dataset distributed on devices: on each device at each time slot, 20\% of on-device data is replaced with new data to simulate the shifting data distributions. We observe that: (1) the static models, both the full model and the compact sub-model, cannot well-adapt to dynamic environments, and the accuracy of the sub-model decreases by around 11\% as data distribution shifts. (2) the updated models could have better accuracy, but updating the model locally on an individual device does not result in satisfactory performance: around 10\% lower than the ideal situation where the sub-model is extracted from the full model updated by new data across devices. 
For inner runtime environment dynamic, Figure \ref{impact_dynamic_env}(b) shows the model inference latency of two models (MobileNetV2~\cite{sandler2018mobilenetv2} and ShuffleNetV2~\cite{ma2018shufflenet}) under different numbers of processes co-running on device. The competition for on-devices resources could significantly add model processing time, increasing up to 5.06$\times$ inference latency with 3 background processes.

With this observation, we are motivated to propose an edge-cloud collaborative learning framework that supports continuous model training and inference on resource-constrained edge devices, allowing their models to keep adapting to dynamic edge environments. 
By collaborating  agile compact models on edge devices and powerful large model on the cloud, the model adaptation could be more effective and efficient compared to the classical cloud-based and the on-device continuous learning methods~\cite{belouadah2019il2m,li2017learning}.
Intuitively, the powerful cloud model can help edge models adapt to the new environment with minimal model re-training overhead by reusing the sub-models for the same environment learned by other edge devices.
The agile edge models can capture new environments efficiently, and transfer this knowledge back to form an updated cloud model for future use. 

\subsection{Design Challenges} 

\begin{figure}
    \centering    
    \includegraphics[width=\linewidth]{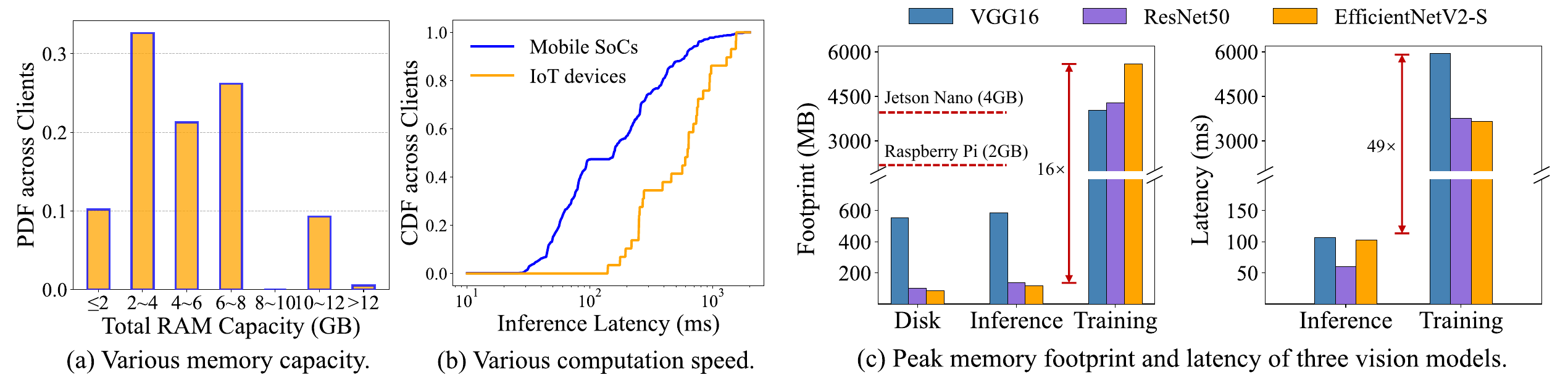}
    % \vspace{-0.8cm}
    \caption{Heterogeneous on-device resources, and intensive resources for on-device model training.}
    % \vspace{-0.3cm}
    \label{heterogeneous devices}
\end{figure}

The design challenges mainly stem from the inherent characteristics of edge devices, \emph{i.e.}, their heterogeneity in data distributions and limitations in system resources. We first analyze these characteristics, and then present design challenges within \emph{personalized edge model derivation} and \emph{heterogeneous edge model aggregation}, respectively. 

Edge devices have strong heterogeneity and large limitations in both systematic and statistical aspects, raising the need for compact and personalized local models. For the systematic aspect, diverse on-device resources (\emph{e.g.}, computation power, memory capacity, and network bandwidth) cause various model performances. %(\emph{e.g.}, more than 50$\times$ the difference in inference latency). 
In Figure \ref{heterogeneous devices}(a) and (b), we showcase the RAM capacity and the inference latency of MobileNetV3~\cite{howard2019searching} in popular mobile phones
%over hundreds of edge devices 
using the statistics from AI Benchmark~\cite{aibenchmark}. We also measured the memory footprint and executing latency of three vision models~\cite{he2016deep,simonyan2014very,tan2021efficientnetv2} on our experimental platform. As shown in Figure~\ref{heterogeneous devices}(c), model training costs more than ten times of peak memory and execution time than model inference, and more than the capacity of Jetson Nano (4GB) and Raspberry Pi (2GB), which hinders edge devices to train a full large model~\cite{dosovitskiy2021an,vaswani2017attention}. 
For the statistical aspect, we observe that 
the local task of an edge device is essentially a sub-task of the global task. For example, in an object recognition application, the global task is to recognize all objects that might be encountered on all edge devices, while the local task of an edge device only needs to recognize a small subset of objects in its surrounding environment~\cite{Zhao2018FederatedLW,li2021hermes}.
The sub-tasks across devices could be quite different, depending on their application contexts, 
and reflected in the non-IID data distributions as well. 

Considering the characteristics of edge devices, we presents two design challenges of ECLM. 

\emph{\textbf{Challenge 1:}} 
Deriving compact and personalized edge models from the large cloud model is non-trivial. It not only needs to derive lightweight edge models with proper structures (to overcome the characteristics in the system aspect), but also needs to derive the models with specialized abilities to deal with target sub-tasks (to overcome the characteristics in the statistical aspect). However, unlike models in some specific areas which have sparsely activated embedding layers that can be decoupled (\emph{e.g.}, recommendation models in~\cite{niu2020billion}), the parameters of large machine learning models are tightly coupled with dense connections~\cite{he2016deep,vaswani2017attention}, making it hard to divide them to form compact sub-models.
%for a certain target sub-task. 
In addition, the frequently changing environments further exacerbate this challenge in that the optimal sub-models for edge devices are changing as well, raising high requirements for low computational complexity of the edge model derivation. 
Although model compression techniques~\cite{wei2016learning,hinton2015distilling,alistarh2017qsgd} such as model pruning~\cite{wei2016learning,han2016deep} and distillation~\cite{hinton2015distilling}, are able to scale down a large cloud model, exhaustively pruning or distilling personalized models for the huge amount of edge devices is prohibitively time-consuming.

\emph{\textbf{Challenge 2:}} 
The personalized edge models are heterogeneous in both model structures and parameters, making it difficult to aggregate them effectively.
We analyze the difficulty in two folds. 
First, 
the commonly used method for transferring knowledge between the models with different  structures is knowledge distillation~\cite{hinton2015distilling,li2019fedmd, lin2020ensemble,itahara2021distillation,cheng2021fedgems}, but it is impractical on edge settings since it may introduce time-consuming model re-training processes and additional computation and storage burdens on edge devices (\emph{e.g.}, calculating model logits on a shared dataset). 
Second, even with the same model structure, parameter conflicts during model aggregation can not be ignored, because the models trained on edge devices with non-IID data distributions could lead to large discrepancies in parameters or gradients. Simply averaging (overlapping) parameters, such as FedAvg~\cite{mcmahan2017communication}, could result in conflicts, which could significantly degrade model performance~\cite{modeling2018ma, crossstitch2016misra}.

It is important to note that these two challenges should not be considered separately, since the way sub-models are derived from the large cloud model determines the sub-model structures and parameters, further affecting the way they are aggregated. Therefore, the two processes should be jointly designed for effective and efficient sub-model derivation and aggregation. In addition, both edge model derivation and aggregation should be lightweight 
for fast model adaptation to frequently changing edge environments. 

\section{ECLM Overview}

\begin{figure}[t]
    \centering    
    \includegraphics[width=\linewidth]{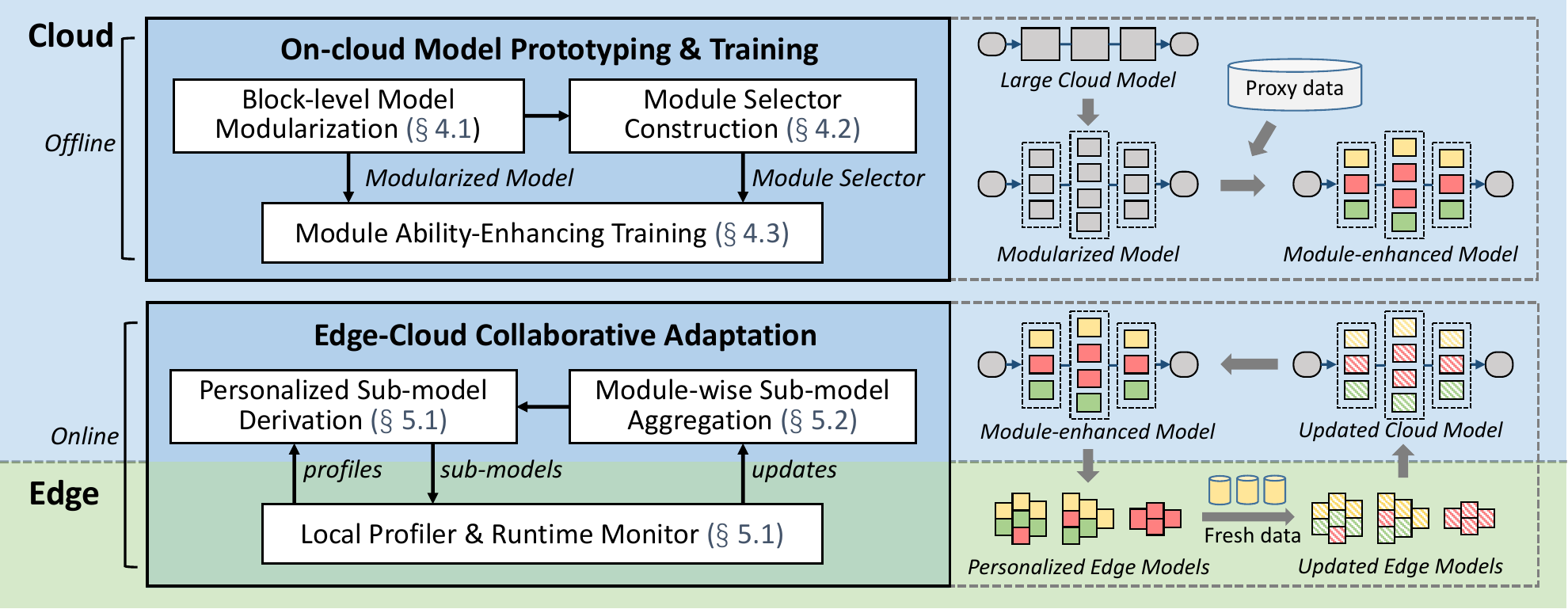}
    % \vspace{-0.7cm}
    \caption{ECLM Overview. In the offline stage, ECLM decomposes the original large cloud model, and trains the modularized model with the corresponding module selector on the cloud. In the online stage, edge devices collaborate with the cloud by periodically obtaining personalized sub-models from the large cloud model, and feeding the updated sub-models back into the cloud model for integrating knowledge learned from massive edge devices.}    
    \label{overview}
    % \vspace{-0.5cm}
\end{figure}

In Figure \textcolor{blue}{\ref{overview}} we illustrate the overall design of ECLM with an offline and an online stage: \emph{on-cloud model prototyping and training} and \emph{edge-cloud collaborative adaptation}, respectively. 

In the on-cloud model prototyping and training stage, we decompose a large cloud model to multiple combinable modules, design a module selector to organize the modules, and prepare the modularized model and the module selector for the subsequent 
%on-device deployment and 
online adaptation stage by jointly training with proxy data on the cloud. 
Specifically, in Block-level Model Modularization component (Section \ref{s4.1}), ECLM takes over an initial large cloud model, identifies the basic blocks within its structure, and then decomposes the cloud model into several module layers, each containing a set of substitute modules. 
In Module Selector Construction component (Section \ref{s4.2}), we construct a unified module selector to organize the modules by intentionally forwarding input samples to the proper modules for processing, which indeed encodes the mapping from sub-tasks to corresponding modules. This ability is learned in the following Module Ability-Enhancing Training process (Section \ref{s4.3}), which decomposes the global task and assign sub-tasks to modules. 
As such, various sub-models with distinct structures and specialized abilities targeting for heterogeneous edge devices can be derived from the large cloud model. 

In the online edge-cloud collaborative adaptation stage, ECLM periodically derives up-to-date personalized sub-models for edge devices, and aggregates the edge model updates to integrate new knowledge for dealing with dynamic edge environments. 
For Personalized Sub-model Derivation (Section \ref{s5.1}), 
An on-device local profiler first characterizes each device's local data distribution and available resources. 
Under on-device resource constraints of each device,
we select the most important modules to form a personalized sub-model with respect to its targeted local task/data distribution. 
During model execution on the edge, devices can either adjust the availability of local modules to flexibly scale their local model sizes for local resource fluctuations, or update their local sub-models with newly collected data to adapt to data distribution shifts. The cloud conducts a Module-wise Sub-model Aggregation (Section \ref{s5.2}) periodically to form an updated cloud model that integrate new knowledge learned by massive edge devices, thus can provide up-to-date sub-models for edge devices in return.

\section{On-cloud Model Prototyping and Training \label{s4}}

\subsection{Block-level Model
Modularization \label{s4.1}}

Instead of directly pre-defining a fixed set of sub-models for edge devices to choose from~\cite{diao2021heterofl,fang2018nestdnn}, our idea is to decompose a large cloud model to multiple reusable modules, which can be selectively  combined to form a sub-model.  
We identify the principle of model modularization in two folds: (i) the modules should form a large design space that is able to derive various personalized sub-models %for heterogeneous edge devices 
at a fine granularity; (ii) each sub-model 
should be responsible for a sub-task, 
\emph{e.g.}, the local task/data distribution on an edge device. The key to defining modules in a large cloud model is to decide module granularity (\emph{i.e.}, the capacity and boundary of an individual module), where neither too small nor too large granularity could fulfill the above principle. 
In this work, we propose \emph{block-level modularization} that identifies basic building blocks within a large model as module layers, and further decomposes each module layer into fine-grained modules. 

% \vspace{3pt}
\noindent\textbf{Identify blocks in a large cloud model.} We identify basic building blocks as the smallest repeated layer patterns within a large cloud model. Each block contains several consecutive network layers, typically beginning with a core layer (\emph{e.g.}, convolutional layers in CNN models), and followed by several auxiliary layers (\emph{e.g.}, BN layers).
For example, a VGG model contains repeated layer sequences such as [Conv, BN, ReLU, Pooling, Dropout], which are identified as VGG blocks, and 
a ResNet block has a similar layer structure but is enhanced with residual connections.
The rationale behind this block definition is that each block is considered to perform a certain function in the learning task, such as feature extraction or classification. Thus, it is reasonable to consider that the sub-model constructed from the connection of these semantic blocks as a whole to undertake a certain sub-task.

\begin{figure}
    \centering    
    \includegraphics[width=0.6\linewidth]{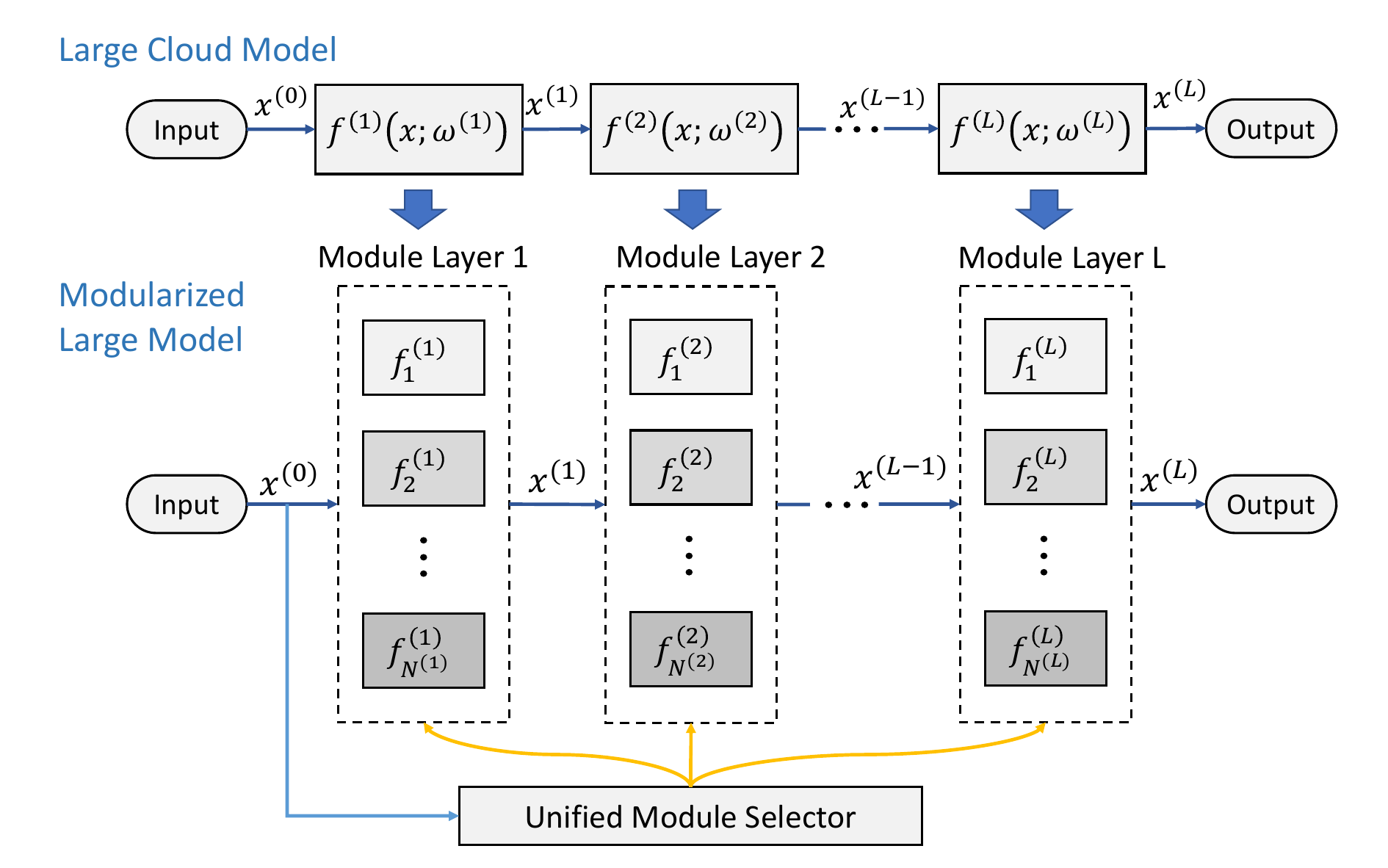}
    % \vspace{-0.7cm}
    \caption{Illustration of modularizing large models.}
    % \vspace{-0.5cm}
    \label{modularization}
\end{figure}

As shown in Figure \ref{modularization}, we formally define the blocks within a large model as functions $f^{(l)}(x^{(l)}; \mathbf{\omega}^{(l)}), l\in\{1,2,\dots, L\}$, where $x^{(l)}$ is the input vector to the $l$th block parameterized by $\mathbf{\omega}^{(l)}$. The output of the $l$th block is fed into the $(l+1)$th block until reaches the final output. As such, a large cloud model $F$ can be represented as the composite of the blocks:

\begin{equation}
    F(x; \mathbf{\omega}) = f^{(L)} \circ f^{(L-1)} \circ \cdots \circ f^{(1)}(x; \mathbf\omega^{(1)}).
\nonumber
\end{equation}

\noindent\textbf{Generate substitute modules for blocks.} To construct a sufficiently large design space, we further generate $N^{(l)}$ substitutable modules $\{f^{(l)}_1, f^{(l)}_2,\dots,f^{(l)}_{N^{(l)}}\}$ for each block $f^{(l)}(x^{(l)}; \mathbf{\omega}^{(l)})$, where a subset of modules can work cooperatively to implement the function of the original block (which is also called module layer thereafter). By doing this, we can have more choices to construct a block and then a sub-model, enabling to generate various sub-models.
%to handle heterogeneous edge devices and sub-tasks. 
Specifically, within a module layer $f^(x^{(l)};\mathbf{\omega}^{(l)})$, each module $i$ is an independent function $f^{(l)}_i (x^{(l)}; \mathbf{\omega}^{(l)}_i)$, and the modules take the same input $x^{(l)}$, but generate distinct outputs. For a given input $x^{(l)}$, we introduce a module selector $\mathbf{g}^{(l)}(x^{(l)}; \mathbf{\theta}^{(l)})$ to selectively activate a subset of its modules in this module layer, and generates the output by combining the outputs of the activated modules. The final output of a module layer is:
\begin{equation}
    f^{(l)}(x^{(l)}; \mathbf{\omega}^{(l)}) = Com_{i \in A}\{f_i^{(l)}(x^{(l)}; \mathbf{\omega}_i^{(l)}); \mathbf{g}^{(l)}(x^{(l)}; \mathbf{\theta}^{(l)})\},
\nonumber
\end{equation}
\noindent where $A$
%= \{a_1, \dots, a_k \}$ 
is the set of activated modules. 

\noindent\textbf{Design network structures for modules.} A module can have arbitrary neural network structures as long as its input and output dimensions are matched with the original block. Without loss of generality, we consider two specific types of modules: shrunk modules and residual modules. A shrunk module $f^{(l)}_i$ adopts the same network layer structure with the original block $f^{(l)}$, but shrinking its size 
by reducing hidden units of its inside network layers, 
such as reducing the number of channels (of convolutional layers) or neurons (of fully connected layers). A residual module provides a residual connection to allow inputs to bypass the current module layer, as it has been demonstrated that not all inputs need layer-by-layer processing for all layers~\cite{he2016deep,wang2018skipnet,laskaridis2020spinn}.

Benefiting from the above module design, ECLM is able to provide a large design space for deriving sub-models.  Take ResNet18 
%in our experiments 
as an example, we modularized it to have 4 module layers, each containing 16 modules. In this way, we can obtain at most $(2^{16})^{4}\approx 2\times10^{19}$ different sub-models with distinct structures and parameters.
%which can be further trained to solve specific sub-tasks. 

\subsection{Module Selector
Construction \label{s4.2}}

In this sub-section, we first consider the module selector for each module layer, 
and then encapsulate the module selectors across all module layers to form a unified module selector.

%\vspace{3pt}
\noindent\textbf{Module selector within a module layer.} 
A module selector $\mathbf{g}^{(l)}$ is responsible for routing inputs $x^{(l)}$ to different subsets of modules in the module layer $l$: $\{f_i^{(l)}| i=1, 2, \cdots, N^{(l)}\}$, which can also be interpreted as a mapping from the sub-tasks to the activated modules. To learn this mapping, we employ a lightweight neural network with a fully connected layer and a softmax layer. The output of the module selector $\mathbf{g}^{(l)}$ for the module layer $l$, 
given an input $x^{(l)}$, is a probability distribution over the modules, which can be regarded as the importance weight of each module with respect to $x^{(l)}$. To reduce on-device computation overhead, a top-$k$ strategy is employed to activate only $k$ out of $N^{(l)}$ available modules for each input $x^{(l)}$.
%, where $k$ could be significantly less than $N^{(l)}$. 
% Next, 
To combine the outputs of the activated modules, we take their weighted summation as the final output of the current module layer, where the weights are the outputs of the module selector. %$\mathbf{g}^{(l)}$. 
That is, the output of a module layer is formulated as:
\begin{equation}
    f(x;\mathbf{\omega}) = \sum_{i \in A} [\mathbf{g}(x; \mathbf{\theta})]_i \cdot f_i(x; \mathbf{\omega}_i),\ A = \text{\emph{Top-k}}(\mathbf{g}(x; \mathbf{\theta})). 
\nonumber
\end{equation}
\noindent\textbf{Unified module selector for all module layers.} 
The selection of the activated modules introduced above is inefficient, because it is a sequential decision-making process: module selector $\mathbf{g}^{(l)}$ takes $x^{(l-1)}$ as the input, which depends on the outputs of all the previous module layers from $1$ to ($l-1$). To speed up this process, we model the module selection for all layers as a one-time decision-making process by combining all $\mathbf{g}^{(l)}$ and an embedding network to form a new neural network, which is called the unified module selector. Specifically, the embedding network extracts intermediate features $h$ from the input $x$ for all $\mathbf{g}^{(l)}$, 
and the output of the unified module selector $\mathbf{g}(x; \mathbf{\theta})$ is the probability distribution $\mathbf{g}^{(l)} (h; \mathbf{\theta}^{(l)}),l\in \{1,\cdots, L\}$ 
%over the activated modules 
for all module layers: 
\begin{equation}
    \mathbf{g}(x; \mathbf{\theta}) = \{\mathbf{g}^{(l)} (h; \mathbf{\theta}^{(l)})|\  h=embed(x;\bar{\mathbf{\theta}}), l\in \{1,\cdots,L\} \}. 
\nonumber
\end{equation}
As such, the unified module selector can determine the activated modules for all module layers at once. We note that this one-time decision-making model is equivalent to the sequential decision-making process when the corresponding neural network is well-designed.

\subsection{End-to-end Model Training  \label{s4.3}}
In this sub-section, we pre-train the modularized large model and the unified module selector on the cloud. 
We propose an end-to-end algorithm for the pre-training process, 
%the module selector and the modularized large model are trained by , 
where the module selector learns to decompose the global task/data distribution into multiple sub-tasks/local data distributions, and maps the sub-tasks to properly activated modules. The activated modules are trained under the coordination of the module selector to deal with the assigned sub-tasks. 

\begin{figure}
    \centering    
    \includegraphics[width=0.7\linewidth]{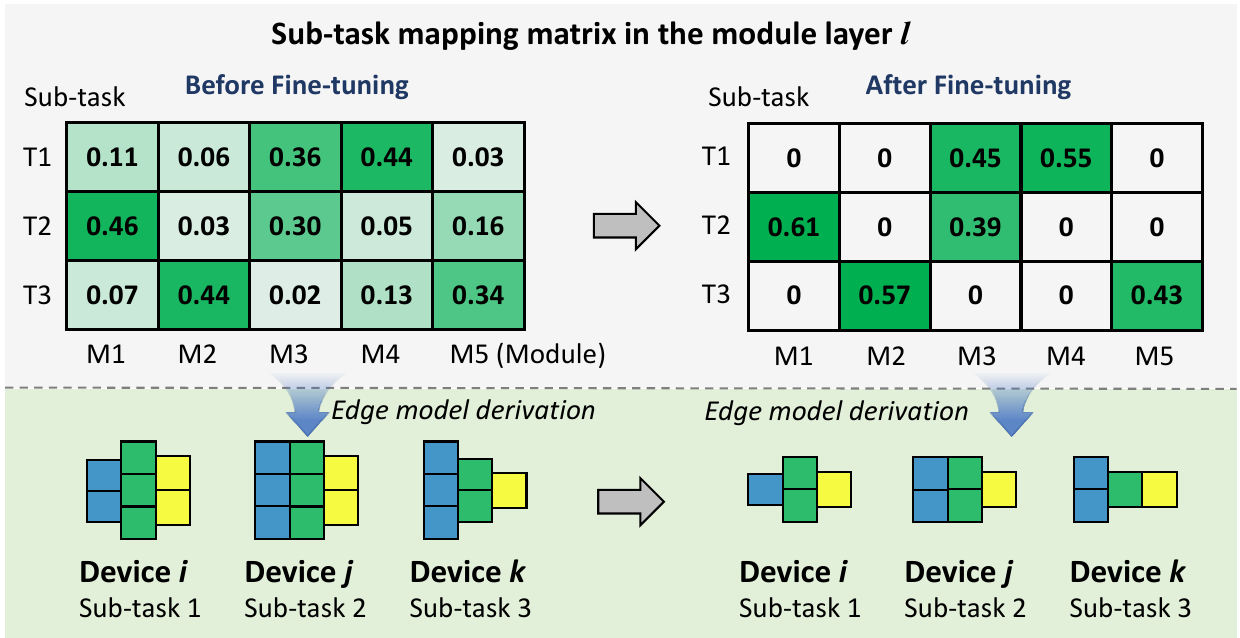}
    \vspace{-0.3cm}
    \caption{Module ability-enhancing training algorithm.}
    \vspace{-0.2cm}
    \label{sub-task mapping matrix}
\end{figure}

\vspace{3pt}
\noindent\textbf{Vanilla end-to-end training.} 
To train such a model, besides the original training loss that aligns model outputs to target labels, we introduce an auxiliary loss term to balance the load of activated modules, where the load is the amount of data samples routed to the associated module\footnote{We employ a noisy top-$k$ technique~\cite{shazeer2017outrageously} to enable end-to-end training with the non-differentiable top-$k$ activating operator.}. 
Without doing this, the modularized model would degrade to a regular model, because the module selector tends to choose the same modules for all data samples (also referred to as module collapse in literature~\cite{kirsch2018modular,shazeer2017outrageously}).
This load-balancing trick can route similar data samples to the same activated modules, and thus a sub-model (a subset of activated modules) can be trained to deal with a specific sub-task/data distribution. Take the classification task as an example, the overall loss function of the end-to-end training is: 
\begin{displaymath}
\mathcal{L}(\hat{\boldsymbol{y}}, \boldsymbol{y}; \hat{\mathbf{g}}) = CrossEntropy(\hat{\boldsymbol{y}}, \boldsymbol{y}) + \lambda \cdot LoadBalance(\hat{\mathbf{g}}), 
\end{displaymath}
where $\hat{\mathbf{g}}$ is the output of the unified module selector and $\lambda$ is  the weight of the load-balancing loss term. 

Although a sub-task decomposition and mapping strategy can be learned automatically by the above end-to-end training, it could be sub-optimal when deriving sub-models for edge devices. This is because the sub-model needed by a given device %to deal with its local sub-task 
might be a combination of a large number of modules, which would break the storage and memory limitation of the edge devices. 
Therefore, we further propose an module-ability enhancing algorithm, where the main goal is to learn a favorable sub-task decomposition and mapping strategy such that each device's local task can be covered by as few modules as possible. 

\vspace{3pt}
\noindent\textbf{Module ability-enhancing training.} 
%The algorithm proceeds in three steps, and 
We use Figure~\ref{sub-task mapping matrix} to illustrate the algorithm with the following three steps: 

(1) \emph{Define application-specific sub-tasks.} We first define the interested sub-tasks with respective to the target application, which is a subset of data samples, having certain common properties, such as the same data distribution, the same class, and etc. The sub-tasks can be defined according to the underlying reasons behind non-IID data distributions across edge devices. 
%such as classes and geographic locations.
For instance, label distribution skew is a common type of non-IID data distributions, where each device only has a small subset of classes, rather than all the classes in the whole dataset. Thus, we can define the classes that usually appear together on a device as a sub-task in this case\footnote{These kinds of information can also be obtained by federated data analysis in a privacy-preserving manner~\cite{elkordy2023federated}.}. In Figure~\ref{sub-task mapping matrix}, we can obtain three sub-tasks and the corresponding sub-task mapping matrix $\boldsymbol{H}_{T \times N}$, where each entry $h_{tn}$ is the normalized total load for module $n$ from all the data samples in sub-task $t$, and can also be interpreted as the probability of mapping sub-task $t$ to module $n$.  

(2) \emph{Identify modules' targeted sub-tasks.} With the current 
$\boldsymbol{H}_{T \times N}$ obtained from the end-to-end training, 
we aim to identify the sub-tasks that a given module $n$ is best at, and let the module focus on these sub-tasks, leaving the other sub-tasks to the other modules. Based on this intuition, we formulate this task identification process as a constrained linear programming problem:

\begin{equation}
\begin{aligned}
\max \ &\ \  \mathbf{H} \odot \mathbf{M}\\
\text{s.t.} &\ \  {\sum_{t=1}^T} \mathbf{M}_{tn} \le \kappa_1, \forall n \in \{1, 2, \dots, N\}, \\
&\ \  {\sum_{n=1}^N} \mathbf{M}_{tn} \le \kappa_2, \forall t \in \{1, 2, \dots, T\}, \\
&\ \  \mathbf{M}_{tn} \in \ \{0, 1\},
\end{aligned}
\end{equation}
where $\mathbf{M}$ is a mask matrix, denoting the  sub-tasks assignment to modules. 
The first  constraint aims to prevent the overload of a given module, that is the load should be less than $\kappa_1$. 
The second constraint limits the maximum number of modules that can be activated by a sub-task.
For objective, we maximize  the element-wise product to preserve the information of the original matrix,
which reflects the strategy learned by the end-to-end training. Preserving this knowledge is conducive to reducing fine-tuning overhead and enhancing convergence speed, since this knowledge embeds the global task's internal structure learned in the end-to-end training stage. 

(3) \emph{Fine-tuning for enhancing modules' abilities.} Based on the obtained target mapping matrix $\mathbf{P} = \mathbf{H} \odot \mathbf{M}$, the goal of the fine-tuning process is two folds: 
one is to train each module using more data from the sub-tasks it focuses on to further enhance its ability on that sub-tasks, 
and the other is to let the module selector update at the guideline of the new sub-task mapping strategy. To this end, the samples from each sub-task are attached by an additional label $\mathbf{g}_{label}$ denoting the recommended modules to activate. The loss function of the fine-tuned training becomes: 
\begin{equation}
\mathcal{L}(\hat{\boldsymbol{y}}, \boldsymbol{y}; \hat{\mathbf{g}}, \mathbf{g}_{label}) = CrossEntropy(\hat{\boldsymbol{y}}, \boldsymbol{y}) + \lambda \cdot KL(\hat{\mathbf{g}}, \mathbf{g}_{label}).
\nonumber
\end{equation}
Following the above process, we could obtain an enhanced modularized cloud model and a unified module selector with a favorable sub-task decomposition and mapping strategy.

\section{Edge-Cloud Collaborative Adaptation \label{s5}}

After obtaining a modularized cloud model and a unified module selector in the model pre-processing stage, we introduce importance-based sub-model derivation to extract personalized sub-models for edge devices with distinct local data distributions and available resources, and module-wise weighted average aggregation to aggregate the updated heterogeneous edge models to form a new cloud model.  

\subsection{
Personalized Sub-model Derivation\label{s5.1}}

To fit personalized sub-models for heterogeneous edge devices within the huge search space of sub-models, ECLM jointly takes local tasks and available on-device system resources into account, achieving flexible tradeoffs between model performance and resource overhead. The objective of fitting sub-models for a given device is to minimize the loss over its local dataset %  $D_i$ $L(D_i; \mathbf{\omega}_j)=\frac{1}{|D_k|}\sum_{k=1}^{|D_k|} l(x_k;\omega_j)$ 
under the resource constraints.
We first define an important metric for modules using the outputs of the unified module selector, and estimate the candidate sub-models' resource overhead with %respective to 
the local resource constraints captured by a local resource profiler. Finally, a set of modules can be chosen to form a sub-model that achieves desired performance-cost tradeoff. 

To identify important modules for edge devices, %As the module selector works on a per-sample basis, 
we define a module's importance score for a given device as the average sample scores of its local data: $Importance(\mathbf{\omega}_i|D_k) = \frac{1}{|D_k|} \sum_{j=1}^{|D_k|} \mathbf{g}(x_j; \mathbf{\theta})_i$, where $D_k$ is the local dataset of device $k$. This importance score embeds the personalized information of the local data distribution, and thus can be used for selecting modules for edge devices. 

To capture resource constraints, we first employ a local resource profiler to capture available resources of edge devices in dynamic runtime environments, %Specifically, the profiler measures available resources in terms of: (i) 
including memory capacity, computational power and network bandwidth. These measurements will serve as the resource constraints in deriving sub-models. 
We next estimate the resource costs of the candidate sub-models on a given device. Since the structure of the modules is determined in the modularization stage, we are able to calculate their resource costs in advance on the cloud.
A sub-model's resource costs are to add up the resource costs of all its containing modules\footnote{The cost estimation method could be extended to advanced methods, \emph{e.g.}, building an additional DNN for cost estimation.}.

After obtaining the importance of modules
%for a given edge device 
and the resource profile, we formulate the personalized sub-model derivation process as a constrained optimization problem:
\begin{equation*}
\begin{aligned}
\max &\ \sum_{i=1}^{|C|} Importance(\mathbf{\omega}_i|D_k) \cdot d_i\\
\text{s.t.} &\ \sum_{i=1}^{|C|} Resource_j(\mathbf{\omega}_i)\cdot d_i \le L_j, j \in \{Comm., Comp., Mem.\}, \\
&\ \ d_i \in \{0, 1\},
\end{aligned}
\end{equation*}
where $C$ denotes the indices of candidate modules. 
To solve this multi-dimensional knapsack problem, 
%in practice, 
we first select the most important module in each module layer to avoid the situation where no module is selected for a certain module layer. Then, the residual problem,  still a multi-dimension knapsack problem, can be solved efficiently using existing optimization tools (\emph{e.g.}, SciPy~\cite{2020SciPy-NMeth} and OR-Tools~\cite{ortools}). As such, we can obtain a set of modules $\mathcal{S}_k=\{\mathbf{\omega}^{(l_1)}_{i_1}, \mathbf{\omega}^{(l_2)}_{i_2}, \cdots, \mathbf{\omega}^{(l_n)}_{i_n}\}$ that forms a personalized sub-model for the edge device. 

In ECLM, edge devices can adjust sub-models on-demand locally to achieve a desired performance-cost tradeoff. 
We allow an edge device to occupy a set of feasible sub-models, which can be  dynamically adjusted to adapt to the runtime resources fluctuation or data distribution shifts. 

\subsection{Module-wise Sub-model Aggregation \label{s5.2}}

The purpose of aggregating updated edge models is to transfer new knowledge learned by edge devices back to the large cloud model, which is important in that: (i) the module are updated continuously with newly collected data. Aggregating the updates back to the large cloud model can further enhance its ability, so as to derive the latest versions of sub-models for the edge devices in return. (ii) aggregating the edge models exploits the updates trained with more data across edge devices, thus the resulting model could be more resistant to overfitting. Besides, the updated modules by some edge devices can be reused for the other edge devices that encounter similar edge environments, thus largely reduce the re-training requirements and overhead. 

To aggregate the heterogeneous edge models, we propose a module-wise weighted average aggregation method. The rationale is that the sub-models are built from the same basic building blocks, \emph{i.e.,} the modules, we can aggregate them in a module-wise manner. 
Specifically, we could update the parameters of module $i$ by calculating the weighted average over the parameters of module $i$ from all sub-models within $\mathbb{U}_i$, which is the set of sub-models that contains module $i$. %Since the modules are activated on a per-sample basis, 
Considering that each module $i$ could be updated a different number of times by different sub-models, %Therefore, 
we exploit the (normalized) importance value of module $i$ with respective to the sub-models as the averaging weights
to balance the contribution of each sub-model. 
That is, the parameters $\mathbf{\omega}_i$ of module $i$ are updated as $\mathbf{\omega}_i^\prime = \sum_{k=1}^{|\mathbb{U}_i|} Importance(\mathbf{\omega}_i|D_k)\cdot \mathbf{\omega}_{ik}^\prime$. This module-wise aggregation reduces the parameter conflicts, because each module is trained by the data samples from
%responsible for 
a specific sub-task without interference from the different sub-tasks on other edge devices. 

\section{Framework and Implementation \label{s6}}

Based on the above discussion, we now can put all the components together to form an edge-cloud collaborative learning framework that keeps adapting models on both the cloud and edge to dynamic edge environments. The overall ECLM framework proceeds as the following four steps: 

(i) In the offline stage, we modularize the large cloud model, and pre-train it and the unified module selector
using the module ability-enhancing training algorithm. Next, the cloud is ready for edge devices to query, either for requesting sub-models or uploading model updates. 

(ii) To request new sub-models, 
each edge device profiles its local data distribution and available system resources.
The local profiles are then used to query the cloud to retrieval  a personalized sub-model.

(iii) During serving on edge devices, sub-models could be updated with the newly collected data locally. When a pre-defined model upload condition is reached, edge devices upload their updated sub-models to the cloud. 
When edge environment changes are detected, edge devices can first adapt their models locally, \emph{e.g.}, adjusting modules from the candidate modules on devices. After on-device adaptation, if the model performance still cannot satisfy the requirements, the edge devices could turn to the cloud for the latest version of sub-modules. 

(iv) After a pre-defined number of module updates are received on the cloud, we aggregate the module updates to transfer the newly learned knowledge on edge devices back to the cloud model, so as to provide the latest versions of modules for edge devices. 

\begin{figure}
    \centering    
    \includegraphics[width=0.55\linewidth]{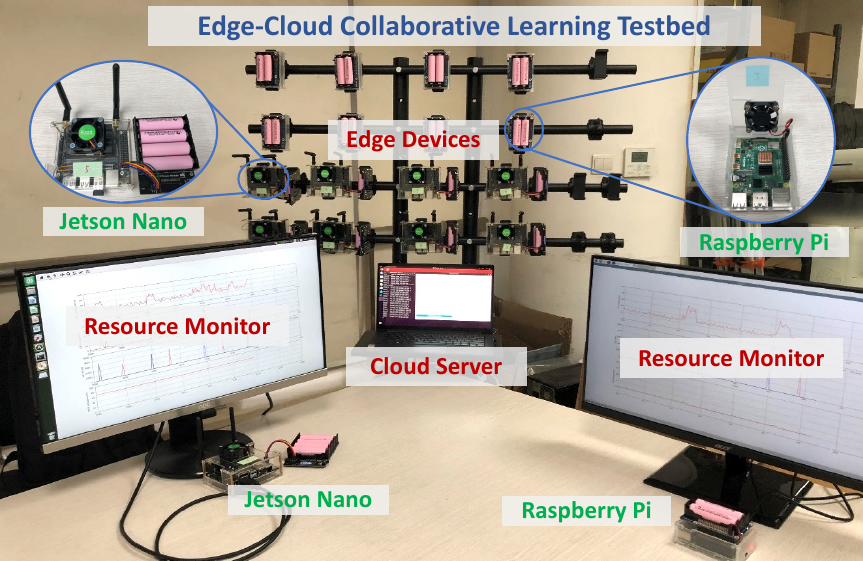} 
    \vspace{-0.2cm}
    \caption{Edge-cloud collaborative learning system testbed with 20 heterogeneous edge devices.}
    \label{implementation}
    \vspace{-0.2cm}
\end{figure}

We implemented ECLM framework  on a simulation platform and a real-world testbed based on PyTorch~\cite{paszke2019pytorch}. Our simulation platform is a Linux server equipped with a 10-core 2.4GHz Intel Xeon Silver 4210R CPU, and two NVIDIA 3090 GPUs. The real-world testbed is shown in Figure \ref{implementation}, which comprises 10 NVIDIA Jetson Nanos and 10 Raspberry Pi 4Bs as edge devices, and a Lenovo laptop as the cloud server. The Nano devices have stronger system performance with on-device GPUs than the Pi devices with CPU only.
%for local computation. 
All devices are equipped with WiFi module, and can connect with the cloud server through a wireless local area network.

\section{Evaluation \label{s7}} 

\subsection{Experimental Methodology}

\noindent\textbf{Tasks, Datasets and Models. }
We evaluate ECLM on three representative applications with four datasets and models to demonstrate its efficiency and effectiveness:

\begin{itemize}
\item  \emph{Mobile Sensing.} Human activity recognition is important %an important task of human-centered computing leveraged by 
for smart devices to understand user behaviors. We use HAR dataset~\cite{Anguita2013apublic} with a 3-layer multi-layer perceptron (MLP) to recognize human activities among 6 categories: walking, walking upstairs, walking downstairs, sitting, standing, and lying-down.
    
\item  \emph{Image Classification.} Image classification is a fundamental task in computer vision. In this task, we use two datasets, CIFAR-10 and CIFAR-100~\cite{krizhevsky2009learning}, with 10 and 100 categories, respectively, %which have different classification difficulties, 
    and employ ResNet18~\cite{he2016deep} and VGG16~\cite{simonyan2014very} models for these two datasets.

\item  \emph{Speech Recognition.} Speech recognition is a basic component of human-computer interaction, where we use Google Speech~\cite{warden2018speech} dataset, and train ResNet34~\cite{he2016deep} to classify audio commands across 35 categories.
\end{itemize}

\noindent\textbf{Data and System heterogeneity.} We consider two common types of non-IID data distributions, \emph{i.e.}, feature skew and label skew. For HAR, we %distribute data according to user IDs: 
assign each device a certain user's data to simulate the %real-world data distribution under the 
feature skew. For the other datasets, we let each device holds only $m$ out of $n$ total classes of data to simulate the label skew. In particular, we test two degrees of data heterogeneity for each dataset (Data Partition 1 and 2) by choosing different values of $m$. Besides, the data volumes across devices are unbalanced, ranging from 50 to 150 samples. 
To simulate real-world hardware heterogeneity on edge devices, we use the statistics from an open-source AI benchmark~\cite{aibenchmark} to sample on-device resource budgets.

\noindent\textbf{Baselines.} We compare ECLM with various baselines in the following paradigms %in the context of 
for dynamic edge environments:

\begin{itemize}

\item \textbf{No Adaptation:} Edge devices use the pre-trained large cloud model without any local adaptation on devices.
    
\item \textbf{On-device Adaptation:} Each edge device adapts its model locally to overcome  dynamic edge environments without any collaboration with the cloud. In this case, we select \emph{Local adaptation} (LA) and \emph{AdaptiveNet} (AN)~\cite{wen2023adaptivenet} as our baselines.
In the LA approach, 
edge devices update their models on devices using newly collected data. In the AN approach, edge devices get a multi-branch model pre-trained on the cloud, and can adjust the model locally to make flexible tradeoffs between model accuracy and inference latency. 
    
\item \textbf{Edge-cloud Collaborative Adaptation:} The edge models on devices collaborate with the cloud model to keep adapting to the dynamic edge environments. In this case, we choose \emph{FedAvg} (FA)~\cite{mcmahan2017communication} and \emph{HeteroFL} (HFL)~\cite{diao2021heterofl} as the baselines. 
\emph{HeteroFL} is a resource-aware federated learning solution, which trains a series of nested models with various sizes for edge devices with different  available resources.
\end{itemize}

\begin{table*}[t]
\definecolor{Gray}{gray}{0.9}
\centering % used for centering table
\renewcommand{\arraystretch}{1.3}
% \small
%\setlength\arrayrulewidth{1.1pt}
\setlength{\tabcolsep}{6pt}
\setlength{\arrayrulewidth}{0.9pt}
\begin{tabularx}{\linewidth}{c @{\hspace{0.3cm}} c c c @{\hspace{0.4cm}} c c c c c c} % centered columns (4 columns)
\hhline{-|-|-|-|-|-|-|-|-|-|} %inserts double horizontal lines
\multirow{2}{*}{Task} & \multirow{2}{*}{Dataset} & \multirow{2}{*}{Model} & \multirow{2}{*}{\makecell{Data\\Per Device}} & \makecell{No\\Adaptation} & \multicolumn{2}{c}{\makecell{On-device\\Adaptation}} & \multicolumn{3}{c}{\makecell{Edge-cloud \\Collaborative Adaptation}} \\ [.6ex] \cline{5-10}  % inserts table 
  & & & & NA & LA & AN & FA & HFL & ECLM \\
%heading
\hhline{-|-|-|-|-|-|-|-|-|-|} % inserts single horizontal line
Sensing & HAR & MLP & 1 subject & 93.96 & 96.07 & 97.42 & 97.35 & 98.31 &  \cellcolor{Gray} \textbf{98.63}
\\ \hhline{-|-|-|-|-|-|-|-|-|-|}
\multirow{4}{*}{\makecell{Image\\Classification}} & \multirow{2}{*}{CIFAR10} & \multirow{2}{*}{ResNet18} & 2 classes & 73.55 & 84.19 & 87.63 & 73.68 & 70.19 & \cellcolor{Gray} \textbf{90.86} \\
\hhline{~~~|-|-|-|-|-|-|-|}
& & & 5 classes & 73.55 & 73.56 & 81.17 & 76.12 & 77.32 & \cellcolor{Gray} \textbf{85.76} \\ 
\hhline{~-|-|-|-|-|-|-|-|-|}
& \multirow{2}{*}{CIFAR100} & \multirow{2}{*}{VGG16} & 10 classes & 56.79 & 67.10  & 69.89 & 60.81 & 52.54 & \cellcolor{Gray} \textbf{74.20} \\
\hhline{~~~-|-|-|-|-|-|-|}
& & & 20 classes & 56.79 & 58.03 & 67.53 & 61.66 & 55.23 & \cellcolor{Gray} \textbf{75.68} \\
\hhline{-|-|-|-|-|-|-|-|-|-|}
\multirow{2}{*}{\makecell{Speech\\Recognition} } & \multirow{2}{*}{\makecell{Google\\Speech}} & \multirow{2}{*}{ResNet34} & 5 classes & 62.72 & 60.52 & 69.33 & 70.48 & 71.73 & \cellcolor{Gray} \textbf{80.87} \\
\hhline{~~~-|-|-|-|-|-|-|}
& & & 10 classes & 62.72 & 59.04 & 67.91 & 73.55 & 72.34 & \cellcolor{Gray} \textbf{77.16} \\
\hhline{-|-|-|-|-|-|-|-|-|-|} %inserts single line
\end{tabularx}
\caption{Model accuracy of ECLM and baselines after an adaptation step.} 
\label{table:e2e-perf}
\vspace{-0.5cm}
\end{table*}

\begin{figure}[t]
    \centering
    \includegraphics[width=\linewidth]{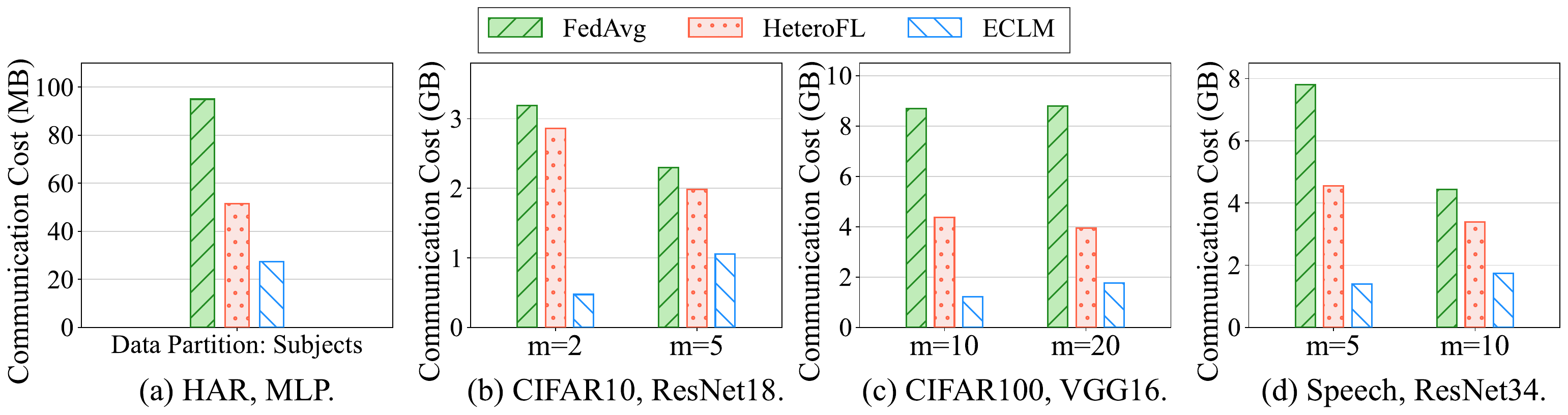}
    \vspace{-0.5cm}
    \caption{Communication costs during adaptation.}
    \label{communication cost}
    \vspace{-0.2cm}
\end{figure}

%\vspace{3pt}
\noindent\textbf{Parameter settings.} %Our experiments use parameter settings as follows. 
For edge-cloud collaborative training, 25 out of 500 devices (5\%) are randomly selected to participate in each communication round. Each participating device trains its model by 3 local epochs. The learning rate is set as 0.001 and the batch size is 16. For on-device adaptation, each edge device fine-tunes the local model for 10 epochs 
%for each adaptation step 
using its local data. 
For model modularization, we employ 1 module layer with 16 modules for MLP model, and 4 module layers each with 16 modules for ResNet18. Since the parameters of VGG16 and ResNet34 are mainly concentrated at the deep layers, we only modularize the last three model blocks as 3 module layers with 32 modules each. 

\subsection{Overall System Performance \label{s7.2}}

To demonstrate the adaptation ability of different approaches in dynamic edge environments, we evaluate the system performance
(\emph{i.e.}, model accuracy and resource costs in terms of communication, memory and latency) after one adaptation step. 
To simulate an adaptation step, we use 30\% of the training dataset as the proxy dataset for model pre-training on the cloud, and the remaining 70\% is distributed to edge devices as newly collected data for adaptation. 

We summarize the model accuracy after the adaptation in Table \textcolor{blue}{\ref{table:e2e-perf}}. The results demonstrate that ECLM outperforms the baselines in all the learning tasks and models. Specifically, 
ECLM has huge superior performance over the No Adaptation approach, indicating the necessity to conduct adaptation. 
Furthermore, ECLM improves model accuracy by 9.06\% and 11.07\% on average compared to on-device adaptation and the other edge-cloud collaborative adaptation methods, respectively. These accuracy improvements are attributed to the efficient collaboration between edge devices and the cloud. 
Compared with the on-device adaptation approaches, 
%Despite the non-IID distribution of data across devices, 
ECLM relies on the large cloud model to flexibly and dynamically derive the personalized sub-model for each edge device. For example, in the speech recognition task, ECLM achieves 80.87\% accuracy, while AN only obtains 69.33\%.  Furthermore, 
compared with the other edge-cloud collaborative adaptation approaches, ECLM effectively aggregates the new sub-models learned from the newly collected data on devices into well-separated modules of the cloud model,
overcoming the problem of non-IID data distributions, which is the major reason to the performance degradation in FA and HFL. 
For example, in CIFAR10 task with $m=2$, ECLM achieves 90.86\% accuracy, significantly outperforming FA (73.68\%) and HFL (70.19\%).

\begin{figure}[t]
    \centering
    \includegraphics[width=0.7\linewidth]{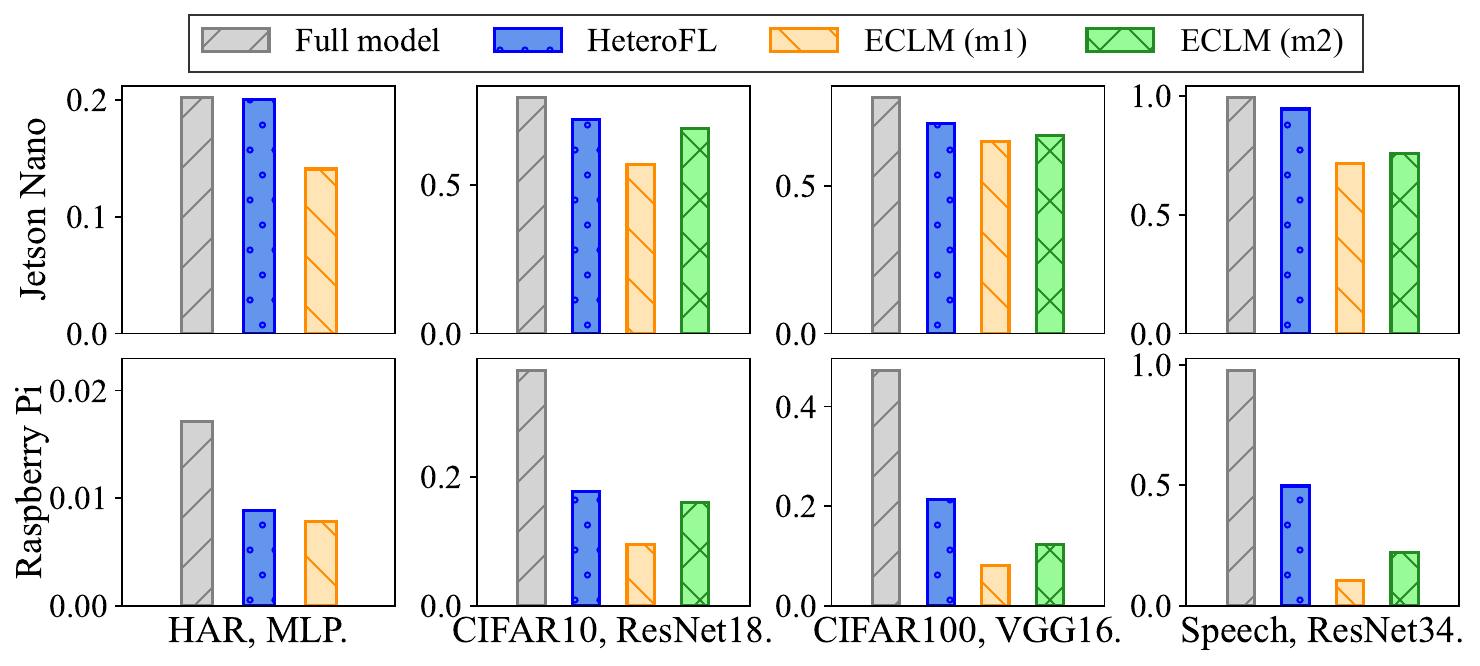}
    \vspace{-0.3cm}
    \caption{Memory footprint (GB) during adaptation.}
    \label{memory footprint}
\end{figure}

\begin{figure}[t]
    \centering
    \includegraphics[width=0.7\linewidth]{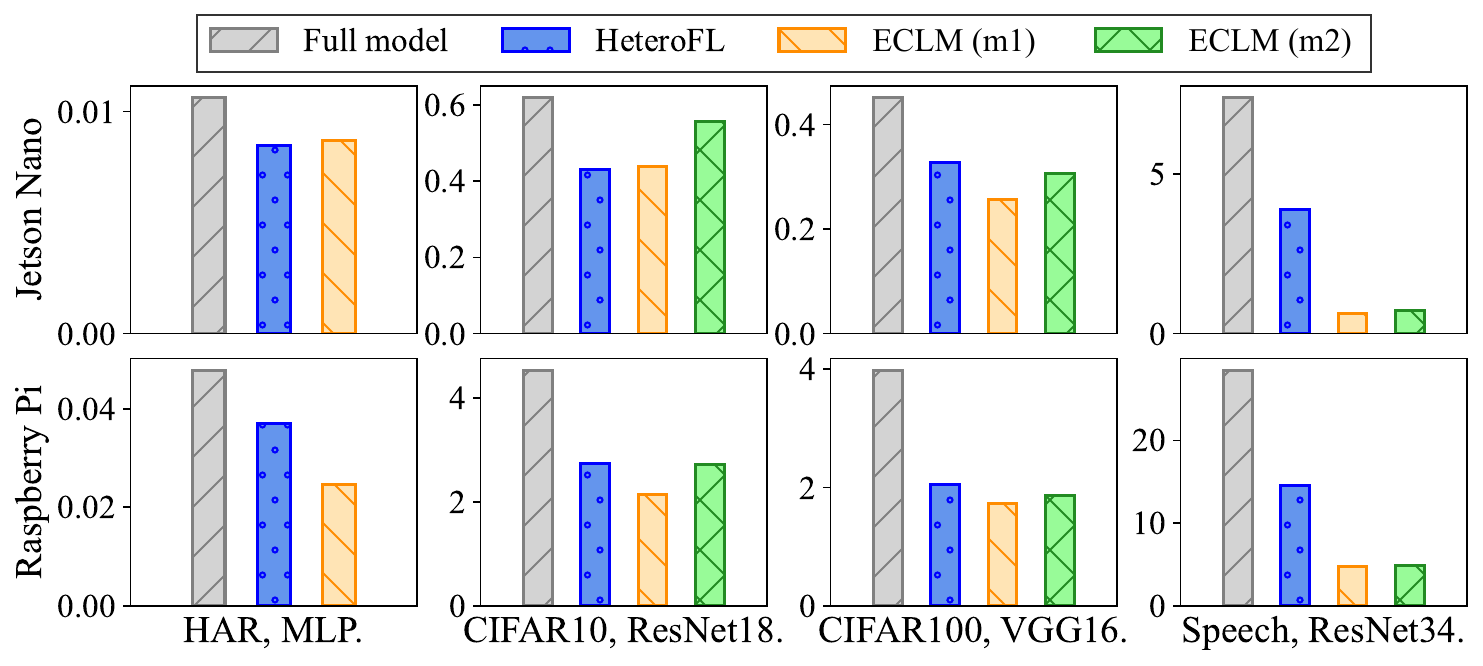}
    \vspace{-0.3cm}
    \caption{Training latency (s) during adaptation.}
    \label{per batch latency}
\end{figure}

We next report the communication costs of the edge-cloud collaborative adaptation strategies 
%during the adaptation process 
in Figure \textcolor{blue}{\ref{communication cost}}. ECLM obtains significant communication cost savings compared to FedAvg and HeteroFL, with average reductions of $4.60\times$ and $2.76\times$, respectively. This is because ECLM only transmits the sub-model parameters, instead of the whole model parameters, between edge devices and the cloud, 
%in each communication round. 
and the size of these sub-models is considerably smaller (\emph{e.g.}, 3.14$\times$ smaller on the speech recognition task) than that of the large cloud model. 
Although HeteroFL also communicates only partial model parameters, %between edge devices and the cloud, 
its lack of consideration for non-IID data distributions leads to a slower convergence time (1.83$\times$ more communication rounds on average than FedAvg).

We now measure the memory footprint in Figure \ref{memory footprint} and per-batch training latency in Figure \ref{per batch latency} on Jetson Nano and Raspberry Pi. Benefiting from the compact sub-models employed by ECLM, we can achieve a remarkable reduction in memory footprint and training latency, with a reduction up to 9.28$\times$ and 11.64$\times$, respectively. Besides, we observe that ECLM demonstrates an even stronger reduction in memory and latency when the cloud model is larger.
This is because 
ECLM can scale down the large model into compact sub-models tailored for edge devices with limited resources. %with lower memory requirements and training latency.

From the above experiment results, we can safely conclude that ECLM has superior performance, not only improving model accuracy after adaptation, but also significantly reducing resource costs for edge devices. 

\begin{figure}[t]
    % \vspace{-0.3cm}
    \centering    
    \includegraphics[width=\linewidth]{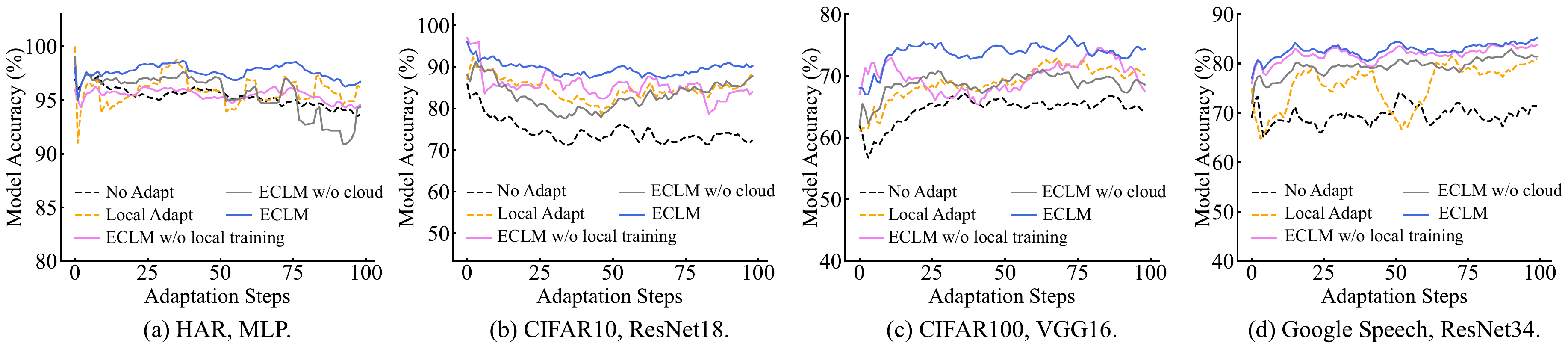}
    \vspace{-0.5cm}
    \caption{Model accuracy after each adaptation step. The first two tasks (HAR and CIFAR10) were performed on  Raspberry Pi, and the other two (CIFAR100 and Speech) were performed on Jetson Nano.} 
    \label{continouous adaptation accuracy.}
    % \vspace{-0.3cm}
\end{figure}

\begin{figure}[t]
    % \vspace{-0.3cm}
    \centering    
    \includegraphics[width=0.7\linewidth]{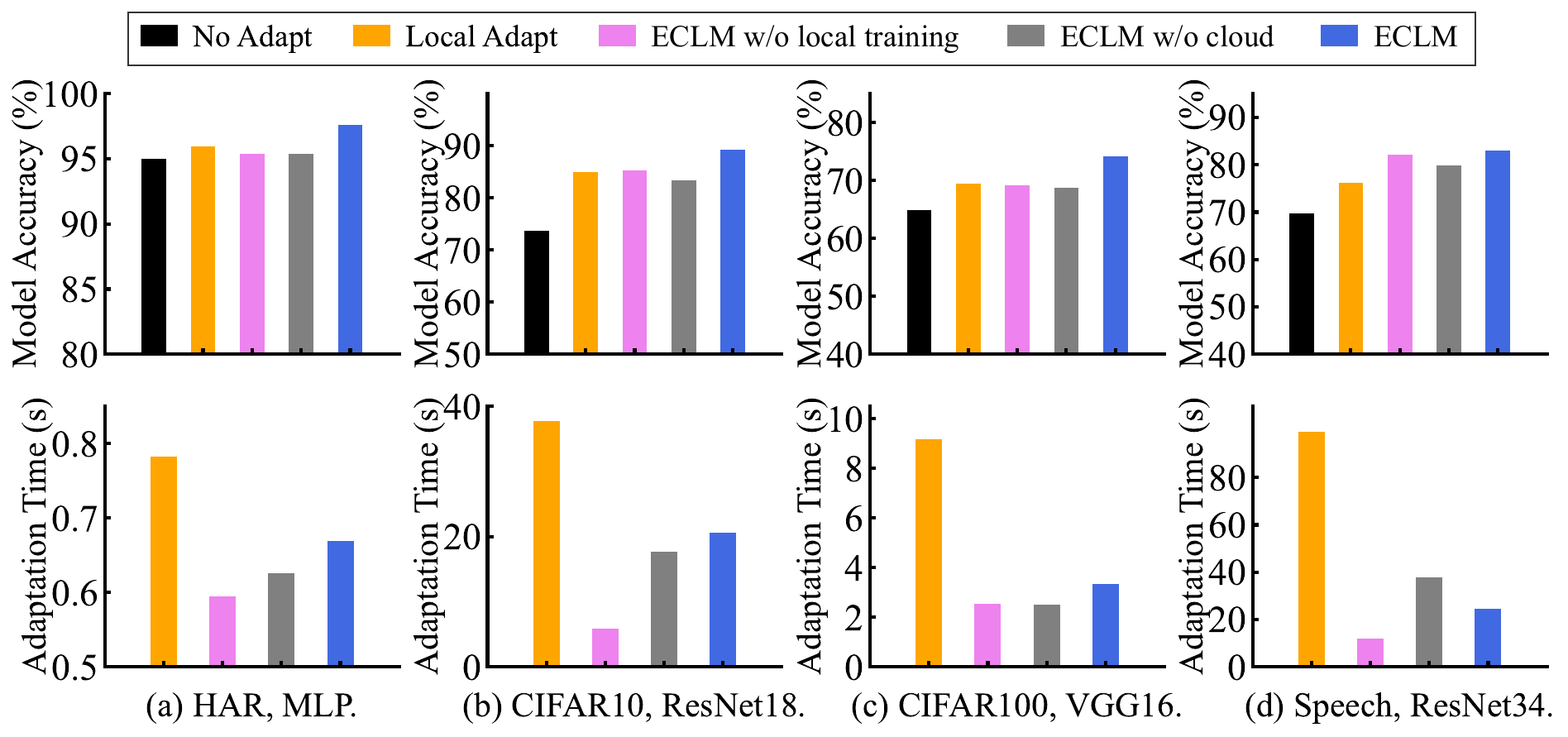}
    \vspace{-0.3cm}
    \caption{Adaptation accuracy and adaptation time.} 
    \label{adaptation accuracy and time}
    % \vspace{-0.4cm}
\end{figure}

\subsection{Continuous Adaptation Performance}
We further evaluate the model adaptation performance on two specific edge devices %(Jetson Nano and Raspberry Pi).
for multiple adaptation steps. %Specifically, we set up 100 time slots on each device. 
In each adaptation step, we randomly replace 50\% of the local data with new data 
to simulate data shifts caused by dynamic edge environments. We also compare two variants of ECLM to provide insights behind its superior performance: (i) ECLM w/o local adaptation: the edge device queries the cloud for a new sub-model in each step without updating the sub-model locally.
(ii) ECLM w/o cloud: the edge device queries the cloud once for a sub-model, and updates it locally without relying on the cloud in the following adaptation steps. 

The model accuracy in each step and the average adaptation accuracy of 100 steps are illustrated in Figure \ref{continouous adaptation accuracy.} and Figure \ref{adaptation accuracy and time}, respectively. ECLM consistently outperforms the baselines, achieving an average improvement in model accuracy of 1.68\%, 4.33\%, 4.72\%, and 6.81\% compared with the LA approach across the four tasks. Again, the advantages of ECLM come from the effective  collaboration between edge devices and the cloud, where the edge devices first exploit the powerful cloud model to obtain a personalized sub-model, and then update the sub-model  with the new data on devices, which can be further used to form a new cloud model. 

We also evaluated the average time cost for each adaptation step in Figure \ref{adaptation accuracy and time}. ECLM outperforms the LA on four tasks, reducing average adaptation times by 14.5\%, 45.5\%, 63.5\%, and 75.3\%, respectively, which demonstrates the efficiency of ECLM in adapting to new environments. The benefits arise from  using a compact sub-model for local training, and the accelerated convergence enabled by the effective edge model aggregation. 

\begin{figure}[t]
    \centering    
    \includegraphics[width=0.85\linewidth]{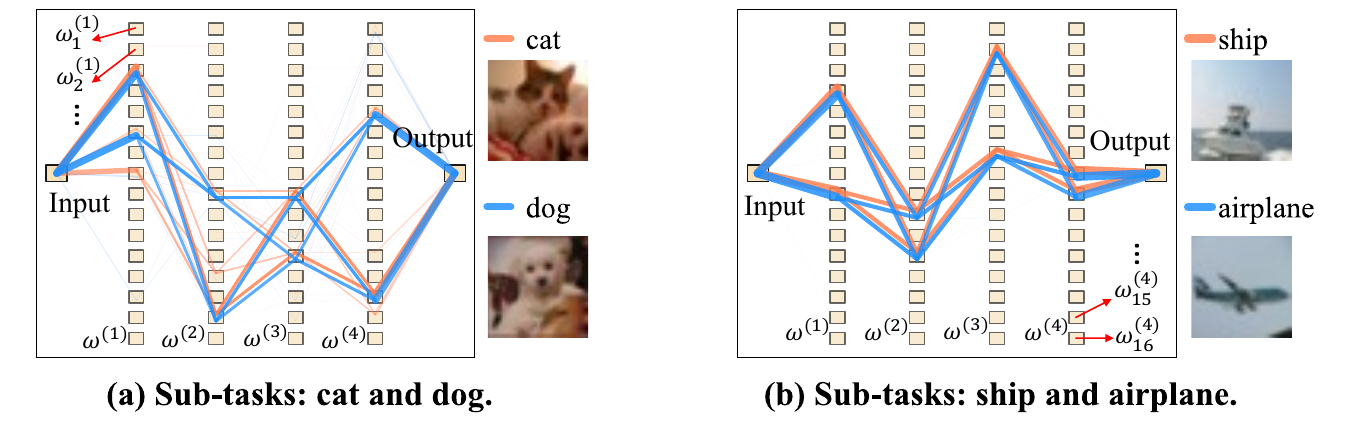}
    \vspace{-0.3cm}
    \caption{Module routing of four sub-tasks (classes) in CIFAR10. Each square represents a module, and the line width is proportional to the amount of data sample passing through that path.}    
    \label{module routing}
    \vspace{-0.3cm}
\end{figure}

\subsection{Performance Breakdown Analysis} 

In this sub-section, we break down ECLM and evaluate its core components to provide insights into the performance of ECLM. 

\noindent{\bf Modularized cloud model and module selector.} To demonstrate the effectiveness of ECLM in sub-task decomposition and mapping, we visualize the activated modules in ResNet18 for 4 sub-tasks in  CIFAR10 dataset in Figure \ref{module routing}, and the results of two similar sub-tasks are presented in the same sub-figure. 
We can observe that the global task is successfully decomposed into sub-tasks in the modularized cloud model, that is different sub-tasks are mapped to distinct sets of activated modules for processing, and similar sub-tasks are mapped to  a similar set of modules. By doing this, ECLM can facilitate parameter sharing among similar sub-tasks, and avoid the conflicts of different sub-tasks in parameter updates, which enhances the overall model performance.
This design also demonstrates the rationale behind deriving compact and personalized sub-models for edge devices.

\begin{figure*}[t]
    % \vspace{-0.1cm}
    \centering    
    \includegraphics[width=\linewidth]{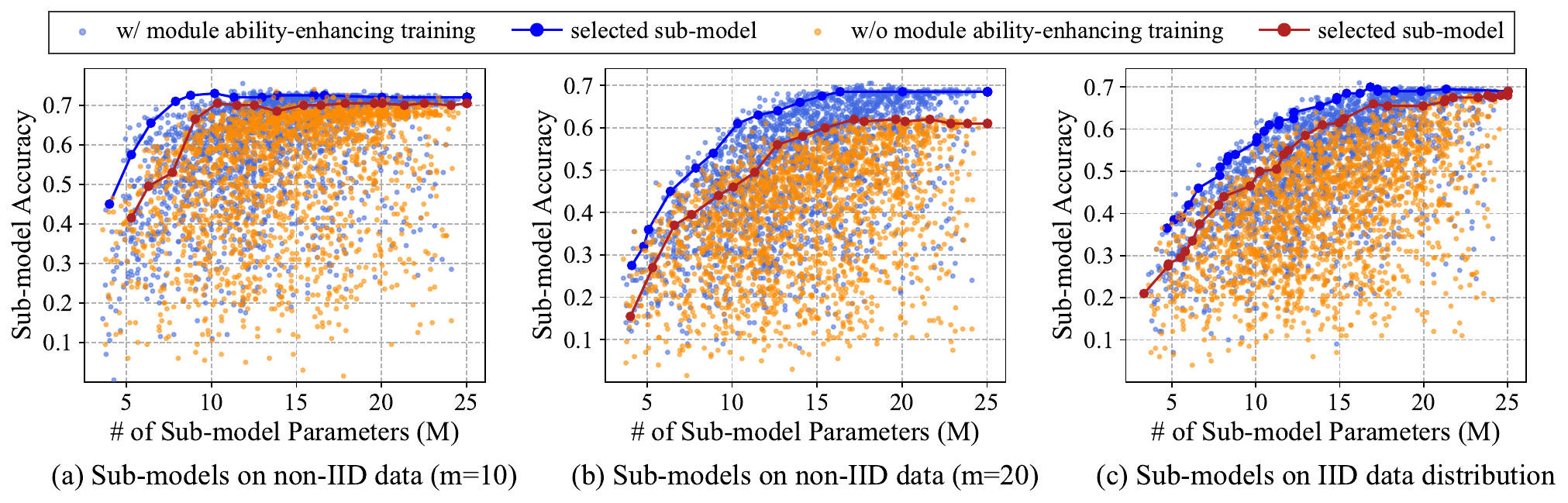}
    \vspace{-0.4cm}
    \caption{Performance of ECLM's sub-models on different non-IID data distributions.}    
    \label{submodel performance}
    % \vspace{-0.4cm}
\end{figure*}

\vspace{2pt}
\noindent{\bf Personalized sub-model derivation.} We utilize VGG16 model trained on CIFAR100 dataset as an example to evaluate the performance of candidate sub-models. As shown in Figure \ref{submodel performance}, each point represents a sub-model generated by randomly selecting a set of modules from each module layer in the modularized cloud model. We have three observations: (i) Our modularized cloud model is able to generate diverse sub-models with varying sizes (from 3M to 25M parameters) and capabilities. (ii) Through our module ability-enhancing training, the performance of the sub-models improves compared to their counterparts of the same size without such training (\emph{e.g.}, the accuracy improves by 11.5\% on average with 5M sub-model parameters). (iii) Our personalized sub-model derivation method effectively identifies near-optimal sub-models under model size constraints, which forms a Pareto optimal curve. 
By comparing with the Pareto optimal curves in Figure \ref{submodel performance},
we can observe that a smaller sub-model is enough to saturate the model performance when m=10, as opposed to the m=20 or 100 (IID).

\begin{figure}[t]
    \centering
    \includegraphics[width=\linewidth]{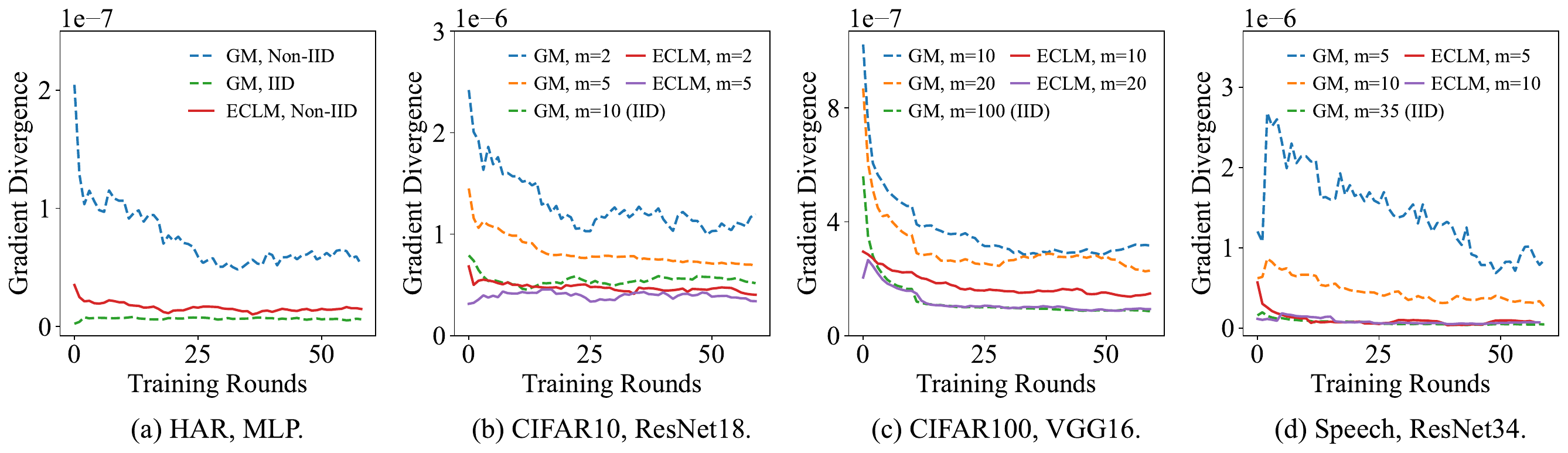}
    \vspace{-0.4cm}
    \caption{Gradient divergence of edge models.}
    \label{gradient divergence}
    % \vspace{-0.4cm}
\end{figure}

\vspace{2pt}
\noindent{\bf Edge model aggregation performance.}% In Section \ref{s7.2}, we have already demonstrated the effectiveness of ECLM in aggregating heterogeneous edge models by the improved model accuracy. 
We now investigate the effectiveness of ECLM in aggregating heterogeneous edge models  
by analyzing the gradient divergence of the updated edge models. Specifically, we record the parameter updates (gradients) of the edge models in each communication round, and calculate their variance as a divergence metric in Figure \ref{gradient divergence}. The gradient divergence increases with the non-IID degrees of local data distributions when directly training a global model (GM) on edge devices. In contrast, with ECLM, the gradient divergence of edge models keeps low (close to training a global model on IID data distributions). 
The benefits stem from our sub-task decomposition, where each module is updated by the data from similar sub-tasks, minimizing interference from other sub-tasks and thus reducing parameter conflicts during edge model aggregation.

\subsection{Sensitivity Analysis}

\begin{figure}[t]
    \centering
    \includegraphics[width=0.9\linewidth]{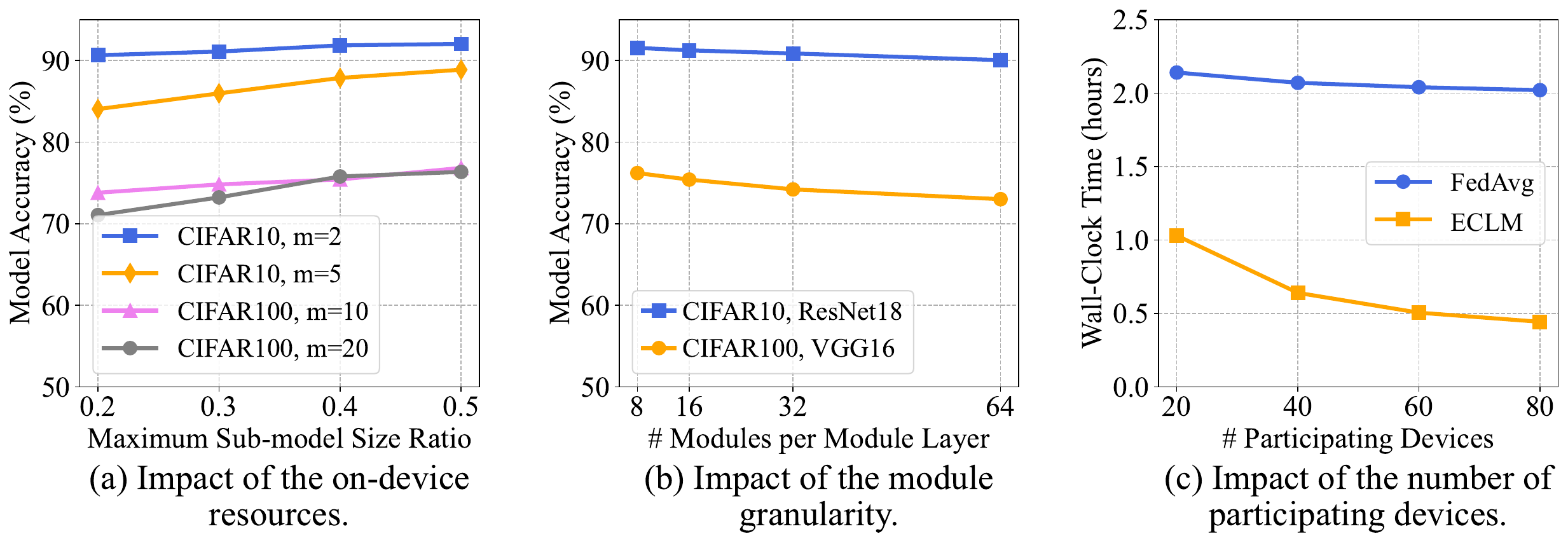}
    \vspace{-0.3cm}
    \caption{Sensitivity Analysis of ECLM.}
    \label{sensitivity analysis}
    \vspace{-0.2cm}
\end{figure}

\noindent{\bf Impact of on-device resources.} To evaluate the robustness of ECLM across various resource capacities, we impose a maximum size constraint on sub-models, and report the model accuracy after adaptation in Figure \ref{sensitivity analysis}(a). 
We observe that increasing the size of sub-models leads to a higher average accuracy, which is expected as larger sub-models contain more modules. %to be trained with new edge data, resulting in improved performance. 
Besides, ECLM shows the ability of edge devices with limited resources: the sub-model with only 20\% size is able to achieve satisfactory performance, and has only a slight accuracy reduction compared to 50\% sub-model (\emph{e.g.}, 3.11\% and 4.19\% for CIFAR10 and CIFAR100, respectively).

\vspace{2pt}
\noindent{\bf Impact of module granularity.} 
We varied the number of modules 
%in each module layer of 
in the cloud model and analyzed the average sub-model accuracy in Figure \ref{sensitivity analysis}(b). 
The results indicate that increasing the number of modules has a slight negative impact on the overall model performance, with reductions of up to 1.5\% and 2.8\% for ResNet18 and VGG16, respectively. This is due to the exponential growth in candidate sub-models when the number of modules increases, which poses challenges for the module selector in learning the sub-task decomposition and mapping strategy. 
Nevertheless, a higher number of candidate modules could allow for finer granularity in adjusting sub-model sizes for heterogeneous devices,
indicating a flexible tradeoff between sub-model sizes and accuracies.

\vspace{2pt}
\noindent{\bf Impact of the number of participating edge devices.} To evaluate  the scalability of ECLM, we measured time-to-accuracy metric with varying numbers of participating devices in each communication round during edge-cloud collaborative learning. We compared these results to the  FedAvg. As shown in Figure \ref{sensitivity analysis}(c), increasing the number of participating devices from 20 to 80 in FedAvg only marginally contributes to training efficiency (speedup by 1.06$\times$), while ECLM consistently has a large training speedup (by 2.33$\times$).
%, showcasing its device-parallel advantage. 
This advantage arises from the modular design of ECLM, which allows each device to update different parts/modules of the large cloud model with minimal interference.

%\clearpage
\section{Related Works \label{related work}}

\noindent\textbf{Dynamic Neural Network.} Unlike traditional static neural networks, dynamic neural networks adaptively activates different parts of the model for different inputs during training and inference, where various methods have been explored
to achieve flexible model scaling 
in terms of model width ~\cite{jocob1991adaptive, shazeer2017outrageously,masoudnia2014mixture,lin2017runtime,wu2018blockdrop}, such as Mixture of Experts (MoE)~\cite{jocob1991adaptive,masoudnia2014mixture}, and depth~\cite{bolukbasi2017adaptive,teerapittayanon2016branchynet,huang2018multiscale,laskaridis2020spinn,wang2018skipnet}, such as SkipNet~\cite{wang2018skipnet}. These approaches mainly focus on adaptive model inference and cannot be directly applied to on-device training due to the new challenges. % in this scenario.
ECLM's modularized model can be considered as a special type of dynamic neural networks, but enhanced with two important aspects: large model decomposition and module ability-enhancing training. 
These designs enable to derive compact sub-models for edge devices for training, further adapting to dynamic edge environments.

% \vspace{3pt}
\vspace{2pt}
\noindent\textbf{On-device model adaptation.} To overcome the drawbacks of static models, previous approaches~\cite{fang2018nestdnn,han2021legodnn,wen2023adaptivenet,yong2016compression,bateni2020neuos} enabled on-device model adaptation to different model structures, \emph{e.g.}, by selecting sub-models from a nested large model which nests multiple DNNs within a single large DNN~\cite{fang2018nestdnn} or by searching suitable sub-models from an offline generated supernet~\cite{wen2023adaptivenet}, to achieve flexible accuracy-latency tradeoffs in facing with changing edge environments. Although effective in resisting resource fluctuations, they do not leverage newly collected data on edge devices, and suffers from performance degradation in dynamic edge environments. 
Instead, ECLM adopts the edge-cloud collaborative learning paradigm, not only allowing edge devices to adapt to resource fluctuation, but also the new local data distribution.

\vspace{2pt}
\noindent\textbf{Edge-cloud collaborative learning.} To fill resources gap between edge devices and the cloud, %and handle heterogeneous edge devices, 
recent studies have explored various approaches to share specific information between the cloud and edge to improve their models, which can be categorized in two types of strategies: \emph{logit sharing-based methods}~\cite{li2019fedmd, lin2020ensemble, itahara2021distillation, he2020group, cheng2021fedgems, Cho2022heterogeneous} and \emph{parameter sharing-based methods}~\cite{caldas2018expanding, Bouacida2021adaptive,horvath2021fjord,diao2021heterofl,li2021hermes,alam2022fedrolex,hong2022efficient,liu2022no,farcas2022model}. The first strategy shares model output logits, and transfers knowledge between models by the knowledge distilllation (KD) technique~\cite{hinton2015distilling}.
In the second strategy, the cloud maintains a large model, from which heterogeneous edge models can be extracted by strategies such as ordered-dropout~\cite{horvath2021fjord} or rolling sub-model extraction~\cite{alam2022fedrolex}.
The sub-models inherit parts of parameters of the large model, and can be aggregated after updating on devices. 
While these methods offer flexibility in defining various model structures for edge devices, they are not designed for dynamic edge environments, due to time-consuming KD and pruning process, or lack of consideration for heterogeneous data distributions on edge devices. 

\section{Conclusion \label{s9}}

In this paper, we have proposed ECLM, an edge-cloud collaborative learning framework for dynamic environments adaptation. In ECLM, edge devices can collaborate with the cloud to quickly adapt to dynamic edge environments,by efficiently deriving personalized sub-models for resource-constrained edge devices, and effectively aggregating updated heterogeneous sub-models back to the cloud model. To enable this, we introduce a modularized large model design and a end-to-end pre-training algorithm to produce the effective edge and cloud models. 
%as the foundation of the edge-cloud collaborative learning pipeline. 
Extensive experiments demonstrate that ECLM not only improves model performance and resource efficiency under dynamic edge environments, but also provides more flexibility for edge devices to do on-device adaptation by module scheduling and updating.

\bibliographystyle{ACM-Reference-Format}
\bibliography{reference}


\begin{thebibliography}{77}

%%% the Web address.

\ifx \showCODEN    \undefined \def \showCODEN     #1{\unskip}     \fi
\ifx \showDOI      \undefined \def \showDOI       #1{#1}\fi
\ifx \showISBNx    \undefined \def \showISBNx     #1{\unskip}     \fi
\ifx \showISBNxiii \undefined \def \showISBNxiii  #1{\unskip}     \fi
\ifx \showISSN     \undefined \def \showISSN      #1{\unskip}     \fi
\ifx \showLCCN     \undefined \def \showLCCN      #1{\unskip}     \fi
\ifx \shownote     \undefined \def \shownote      #1{#1}          \fi
\ifx \showarticletitle \undefined \def \showarticletitle #1{#1}   \fi
\ifx \showURL      \undefined \def \showURL       {\relax}        \fi
% invisible to TeX
\providecommand\bibfield[2]{#2}
\providecommand\bibinfo[2]{#2}
\providecommand\natexlab[1]{#1}
\providecommand\showeprint[2][]{arXiv:#2}

\bibitem[aib(2022)]%
        {aibenchmark}
 \bibinfo{year}{2022}\natexlab{}.
\newblock \bibinfo{title}{AI Benchmark: All About Deep Learning on Smart phones}.
\newblock
\newblock
\urldef\tempurl%
\url{http://ai-benchmark.com/ranking_ deeplearning_detailed.html}
\showURL{%
\tempurl}


\bibitem[Alam et~al\mbox{.}(2022)]%
        {alam2022fedrolex}
\bibfield{author}{\bibinfo{person}{Samiul Alam}, \bibinfo{person}{Luyang Liu}, \bibinfo{person}{Ming Yan}, {and} \bibinfo{person}{Mi Zhang}.} \bibinfo{year}{2022}\natexlab{}.
\newblock \showarticletitle{FedRolex: Model-Heterogeneous Federated Learning with Rolling Sub-Model Extraction}. In \bibinfo{booktitle}{\emph{Advances in Neural Information Processing Systems (NeurIPS)}}.
\newblock


\bibitem[Alistarh et~al\mbox{.}(2017)]%
        {alistarh2017qsgd}
\bibfield{author}{\bibinfo{person}{Dan Alistarh}, \bibinfo{person}{Demjan Grubic}, \bibinfo{person}{Jerry Li}, \bibinfo{person}{Ryota Tomioka}, {and} \bibinfo{person}{Milan Vojnovic}.} \bibinfo{year}{2017}\natexlab{}.
\newblock \showarticletitle{QSGD: Communication-efficient SGD via gradient quantization and encoding}.
\newblock \bibinfo{journal}{\emph{Advances in neural information processing systems (NeurIPS)}}.
\newblock


\bibitem[Anguita et~al\mbox{.}(2013)]%
        {Anguita2013apublic}
\bibfield{author}{\bibinfo{person}{Davide Anguita}, \bibinfo{person}{Alessandro Ghio}, \bibinfo{person}{Luca Oneto}, \bibinfo{person}{Xavier Parra~Perez}, {and} \bibinfo{person}{Jorge~Luis Reyes~Ortiz}.} \bibinfo{year}{2013}\natexlab{}.
\newblock \showarticletitle{A public domain dataset for human activity recognition using smartphones}. In \bibinfo{booktitle}{\emph{Proceedings of the 21th international European symposium on artificial neural networks, computational intelligence and machine learning (ESANN)}}.
\newblock


\bibitem[Bateni and Liu(2020)]%
        {bateni2020neuos}
\bibfield{author}{\bibinfo{person}{Soroush Bateni} {and} \bibinfo{person}{Cong Liu}.} \bibinfo{year}{2020}\natexlab{}.
\newblock \showarticletitle{NeuOS: A Latency-Predictable Multi-Dimensional Optimization Framework for DNN-Driven Autonomous Systems}. In \bibinfo{booktitle}{\emph{Proceedings of the 2020 USENIX Conference on Usenix Annual Technical Conference (USENIX ATC)}}.
\newblock


\bibitem[Belouadah and Popescu(2019)]%
        {belouadah2019il2m}
\bibfield{author}{\bibinfo{person}{Eden Belouadah} {and} \bibinfo{person}{Adrian Popescu}.} \bibinfo{year}{2019}\natexlab{}.
\newblock \showarticletitle{Il2m: Class incremental learning with dual memory}. In \bibinfo{booktitle}{\emph{Proceedings of the IEEE/CVF international conference on computer vision (CVPR)}}.
\newblock


\bibitem[Bhardwaj et~al\mbox{.}(2022)]%
        {bhardwaj2022ekya}
\bibfield{author}{\bibinfo{person}{Romil Bhardwaj}, \bibinfo{person}{Zhengxu Xia}, \bibinfo{person}{Ganesh Ananthanarayanan}, \bibinfo{person}{Junchen Jiang}, \bibinfo{person}{Yuanchao Shu}, \bibinfo{person}{Nikolaos Karianakis}, \bibinfo{person}{Kevin Hsieh}, \bibinfo{person}{Paramvir Bahl}, {and} \bibinfo{person}{Ion Stoica}.} \bibinfo{year}{2022}\natexlab{}.
\newblock \showarticletitle{Ekya: Continuous learning of video analytics models on edge compute servers}. In \bibinfo{booktitle}{\emph{19th USENIX Symposium on Networked Systems Design and Implementation (NSDI 2022)}}.
\newblock


\bibitem[Bolukbasi et~al\mbox{.}(2017)]%
        {bolukbasi2017adaptive}
\bibfield{author}{\bibinfo{person}{Tolga Bolukbasi}, \bibinfo{person}{Joseph Wang}, \bibinfo{person}{Ofer Dekel}, {and} \bibinfo{person}{Venkatesh Saligrama}.} \bibinfo{year}{2017}\natexlab{}.
\newblock \showarticletitle{Adaptive neural networks for efficient inference}. In \bibinfo{booktitle}{\emph{International Conference on Machine Learning (ICML)}}. PMLR.
\newblock


\bibitem[Bouacida et~al\mbox{.}(2021)]%
        {Bouacida2021adaptive}
\bibfield{author}{\bibinfo{person}{Nader Bouacida}, \bibinfo{person}{Jiahui Hou}, \bibinfo{person}{Hui Zang}, {and} \bibinfo{person}{Xin Liu}.} \bibinfo{year}{2021}\natexlab{}.
\newblock \showarticletitle{Adaptive Federated Dropout: Improving Communication Efficiency and Generalization for Federated Learning}. In \bibinfo{booktitle}{\emph{IEEE INFOCOM 2021 - IEEE Conference on Computer Communications Workshops (INFOCOM WKSHPS)}}.
\newblock


\bibitem[Caldas et~al\mbox{.}(2018)]%
        {caldas2018expanding}
\bibfield{author}{\bibinfo{person}{Sebastian Caldas}, \bibinfo{person}{Jakub Kone{\v{c}}ny}, \bibinfo{person}{H~Brendan McMahan}, {and} \bibinfo{person}{Ameet Talwalkar}.} \bibinfo{year}{2018}\natexlab{}.
\newblock \showarticletitle{Expanding the reach of federated learning by reducing client resource requirements}.
\newblock \bibinfo{journal}{\emph{arXiv preprint arXiv:1812.07210}} (\bibinfo{year}{2018}).
\newblock


\bibitem[Cheng et~al\mbox{.}(2021)]%
        {cheng2021fedgems}
\bibfield{author}{\bibinfo{person}{Sijie Cheng}, \bibinfo{person}{Jingwen Wu}, \bibinfo{person}{Yanghua Xiao}, {and} \bibinfo{person}{Yang Liu}.} \bibinfo{year}{2021}\natexlab{}.
\newblock \showarticletitle{Fedgems: Federated learning of larger server models via selective knowledge fusion}.
\newblock \bibinfo{journal}{\emph{arXiv preprint arXiv:2110.11027}} (\bibinfo{year}{2021}).
\newblock


\bibitem[Cho et~al\mbox{.}(2022)]%
        {Cho2022heterogeneous}
\bibfield{author}{\bibinfo{person}{Yae~Jee Cho}, \bibinfo{person}{Andre Manoel}, \bibinfo{person}{Gauri Joshi}, \bibinfo{person}{Robert Sim}, {and} \bibinfo{person}{Dimitrios Dimitriadis}.} \bibinfo{year}{2022}\natexlab{}.
\newblock \showarticletitle{Heterogeneous Ensemble Knowledge Transfer for Training Large Models in Federated Learning}. In \bibinfo{booktitle}{\emph{Proceedings of the Thirty-First International Joint Conference on Artificial Intelligence (IJCAI)}}.
\newblock


\bibitem[Diao et~al\mbox{.}(2021)]%
        {diao2021heterofl}
\bibfield{author}{\bibinfo{person}{Enmao Diao}, \bibinfo{person}{Jie Ding}, {and} \bibinfo{person}{Vahid Tarokh}.} \bibinfo{year}{2021}\natexlab{}.
\newblock \showarticletitle{HeteroFL: Computation and Communication Efficient Federated Learning for Heterogeneous Clients}. In \bibinfo{booktitle}{\emph{International Conference on Learning Representations (ICLR)}}.
\newblock


\bibitem[Dosovitskiy et~al\mbox{.}(2021)]%
        {dosovitskiy2021an}
\bibfield{author}{\bibinfo{person}{Alexey Dosovitskiy}, \bibinfo{person}{Lucas Beyer}, \bibinfo{person}{Alexander Kolesnikov}, \bibinfo{person}{Dirk Weissenborn}, \bibinfo{person}{Xiaohua Zhai}, \bibinfo{person}{Thomas Unterthiner}, \bibinfo{person}{Mostafa Dehghani}, \bibinfo{person}{Matthias Minderer}, \bibinfo{person}{Georg Heigold}, \bibinfo{person}{Sylvain Gelly}, \bibinfo{person}{Jakob Uszkoreit}, {and} \bibinfo{person}{Neil Houlsby}.} \bibinfo{year}{2021}\natexlab{}.
\newblock \showarticletitle{An Image is Worth 16x16 Words: Transformers for Image Recognition at Scale}. In \bibinfo{booktitle}{\emph{International Conference on Learning Representations (ICLR)}}.
\newblock


\bibitem[Elkordy et~al\mbox{.}(2023)]%
        {elkordy2023federated}
\bibfield{author}{\bibinfo{person}{Ahmed~Roushdy Elkordy}, \bibinfo{person}{Yahya~H Ezzeldin}, \bibinfo{person}{Shanshan Han}, \bibinfo{person}{Shantanu Sharma}, \bibinfo{person}{Chaoyang He}, \bibinfo{person}{Sharad Mehrotra}, \bibinfo{person}{Salman Avestimehr}, {et~al\mbox{.}}} \bibinfo{year}{2023}\natexlab{}.
\newblock \showarticletitle{Federated analytics: A survey}.
\newblock \bibinfo{journal}{\emph{APSIPA Transactions on Signal and Information Processing}} \bibinfo{volume}{12}, \bibinfo{number}{1} (\bibinfo{year}{2023}).
\newblock


\bibitem[Fang et~al\mbox{.}(2018)]%
        {fang2018nestdnn}
\bibfield{author}{\bibinfo{person}{Biyi Fang}, \bibinfo{person}{Xiao Zeng}, {and} \bibinfo{person}{Mi Zhang}.} \bibinfo{year}{2018}\natexlab{}.
\newblock \showarticletitle{NestDNN: Resource-Aware Multi-Tenant On-Device Deep Learning for Continuous Mobile Vision}. In \bibinfo{booktitle}{\emph{Proceedings of the 24th Annual International Conference on Mobile Computing and Networking (MobiCom)}}.
\newblock


\bibitem[Farcas et~al\mbox{.}(2022)]%
        {farcas2022model}
\bibfield{author}{\bibinfo{person}{Allen-Jasmin Farcas}, \bibinfo{person}{Xiaohan Chen}, \bibinfo{person}{Zhangyang Wang}, {and} \bibinfo{person}{Radu Marculescu}.} \bibinfo{year}{2022}\natexlab{}.
\newblock \showarticletitle{Model elasticity for hardware heterogeneity in federated learning systems}. In \bibinfo{booktitle}{\emph{Proceedings of the 1st ACM Workshop on Data Privacy and Federated Learning Technologies for Mobile Edge Network (MobiCom FedEdge)}}.
\newblock


\bibitem[{Google LLC}(2021)]%
        {ortools}
\bibfield{author}{\bibinfo{person}{{Google LLC}}.} \bibinfo{year}{2021}\natexlab{}.
\newblock \bibinfo{title}{{OR-Tools}: Operations Research Tools}.
\newblock \bibinfo{howpublished}{\url{https://developers.google.com/optimization}}.
\newblock


\bibitem[Gupta and Raskar(2018)]%
        {Gupta2018DistributedLO}
\bibfield{author}{\bibinfo{person}{Otkrist Gupta} {and} \bibinfo{person}{Ramesh Raskar}.} \bibinfo{year}{2018}\natexlab{}.
\newblock \showarticletitle{Distributed learning of deep neural network over multiple agents}.
\newblock \bibinfo{journal}{\emph{J. Netw. Comput. Appl. (JNCA)}}  \bibinfo{volume}{116} (\bibinfo{year}{2018}).
\newblock


\bibitem[Han et~al\mbox{.}(2021)]%
        {han2021legodnn}
\bibfield{author}{\bibinfo{person}{Rui Han}, \bibinfo{person}{Qinglong Zhang}, \bibinfo{person}{Chi~Harold Liu}, \bibinfo{person}{Guoren Wang}, \bibinfo{person}{Jian Tang}, {and} \bibinfo{person}{Lydia~Y. Chen}.} \bibinfo{year}{2021}\natexlab{}.
\newblock \showarticletitle{LegoDNN: Block-Grained Scaling of Deep Neural Networks for Mobile Vision}. In \bibinfo{booktitle}{\emph{Proceedings of the 27th Annual International Conference on Mobile Computing and Networking (MobiCom)}}.
\newblock


\bibitem[Han et~al\mbox{.}(2015)]%
        {han2016deep}
\bibfield{author}{\bibinfo{person}{Song Han}, \bibinfo{person}{Huizi Mao}, {and} \bibinfo{person}{William~J. Dally}.} \bibinfo{year}{2015}\natexlab{}.
\newblock \showarticletitle{Deep Compression: Compressing Deep Neural Network with Pruning, Trained Quantization and Huffman Coding}. In \bibinfo{booktitle}{\emph{International Conference on Learning Representations (ICLR)}}.
\newblock


\bibitem[He et~al\mbox{.}(2020)]%
        {he2020group}
\bibfield{author}{\bibinfo{person}{Chaoyang He}, \bibinfo{person}{Murali Annavaram}, {and} \bibinfo{person}{Salman Avestimehr}.} \bibinfo{year}{2020}\natexlab{}.
\newblock \showarticletitle{Group Knowledge Transfer: Federated Learning of Large CNNs at the Edge}. In \bibinfo{booktitle}{\emph{Proceedings of the 34th International Conference on Neural Information Processing Systems (NeurIPS)}}.
\newblock


\bibitem[He et~al\mbox{.}(2016)]%
        {he2016deep}
\bibfield{author}{\bibinfo{person}{Kaiming He}, \bibinfo{person}{Xiangyu Zhang}, \bibinfo{person}{Shaoqing Ren}, {and} \bibinfo{person}{Jian Sun}.} \bibinfo{year}{2016}\natexlab{}.
\newblock \showarticletitle{Deep residual learning for image recognition}. In \bibinfo{booktitle}{\emph{Proceedings of the IEEE conference on computer vision and pattern recognition (CVPR)}}.
\newblock


\bibitem[Hinton et~al\mbox{.}(2015)]%
        {hinton2015distilling}
\bibfield{author}{\bibinfo{person}{Geoffrey Hinton}, \bibinfo{person}{Oriol Vinyals}, {and} \bibinfo{person}{Jeff Dean}.} \bibinfo{year}{2015}\natexlab{}.
\newblock \showarticletitle{Distilling the knowledge in a neural network}.
\newblock \bibinfo{journal}{\emph{arXiv preprint arXiv:1503.02531}} (\bibinfo{year}{2015}).
\newblock


\bibitem[Hong et~al\mbox{.}(2022)]%
        {hong2022efficient}
\bibfield{author}{\bibinfo{person}{Junyuan Hong}, \bibinfo{person}{Haotao Wang}, \bibinfo{person}{Zhangyang Wang}, {and} \bibinfo{person}{Jiayu Zhou}.} \bibinfo{year}{2022}\natexlab{}.
\newblock \showarticletitle{Efficient Split-Mix Federated Learning for On-Demand and In-Situ Customization}. In \bibinfo{booktitle}{\emph{International Conference on Learning Representations (ICLR)}}.
\newblock


\bibitem[Horvath et~al\mbox{.}(2021)]%
        {horvath2021fjord}
\bibfield{author}{\bibinfo{person}{Samuel Horvath}, \bibinfo{person}{Stefanos Laskaridis}, \bibinfo{person}{Mario Almeida}, \bibinfo{person}{Ilias Leontiadis}, \bibinfo{person}{Stylianos Venieris}, {and} \bibinfo{person}{Nicholas~Donald Lane}.} \bibinfo{year}{2021}\natexlab{}.
\newblock \showarticletitle{FjORD: Fair and Accurate Federated Learning under heterogeneous targets with Ordered Dropout}. In \bibinfo{booktitle}{\emph{Proceedings of the 35th International Conference on Neural Information Processing Systems (NeurIPS)}}.
\newblock


\bibitem[Howard et~al\mbox{.}(2019)]%
        {howard2019searching}
\bibfield{author}{\bibinfo{person}{Andrew Howard}, \bibinfo{person}{Mark Sandler}, \bibinfo{person}{Grace Chu}, \bibinfo{person}{Liang-Chieh Chen}, \bibinfo{person}{Bo Chen}, \bibinfo{person}{Mingxing Tan}, \bibinfo{person}{Weijun Wang}, \bibinfo{person}{Yukun Zhu}, \bibinfo{person}{Ruoming Pang}, \bibinfo{person}{Vijay Vasudevan}, {et~al\mbox{.}}} \bibinfo{year}{2019}\natexlab{}.
\newblock \showarticletitle{Searching for mobilenetv3}. In \bibinfo{booktitle}{\emph{Proceedings of the IEEE/CVF international conference on computer vision (ICCV)}}.
\newblock


\bibitem[Huang et~al\mbox{.}(2018)]%
        {huang2018multiscale}
\bibfield{author}{\bibinfo{person}{Gao Huang}, \bibinfo{person}{Danlu Chen}, \bibinfo{person}{Tianhong Li}, \bibinfo{person}{Felix Wu}, \bibinfo{person}{Laurens van~der Maaten}, {and} \bibinfo{person}{Kilian Weinberger}.} \bibinfo{year}{2018}\natexlab{}.
\newblock \showarticletitle{Multi-Scale Dense Networks for Resource Efficient Image Classification}. In \bibinfo{booktitle}{\emph{International Conference on Learning Representations (ICLR)}}.
\newblock


\bibitem[Itahara et~al\mbox{.}(2021)]%
        {itahara2021distillation}
\bibfield{author}{\bibinfo{person}{Sohei Itahara}, \bibinfo{person}{Takayuki Nishio}, \bibinfo{person}{Yusuke Koda}, \bibinfo{person}{Masahiro Morikura}, {and} \bibinfo{person}{Koji Yamamoto}.} \bibinfo{year}{2021}\natexlab{}.
\newblock \showarticletitle{Distillation-based semi-supervised federated learning for communication-efficient collaborative training with non-iid private data}.
\newblock \bibinfo{journal}{\emph{IEEE Transactions on Mobile Computing (TMC)}} \bibinfo{volume}{22}, \bibinfo{number}{1} (\bibinfo{year}{2021}).
\newblock


\bibitem[Jacobs et~al\mbox{.}(1991)]%
        {jocob1991adaptive}
\bibfield{author}{\bibinfo{person}{Robert~A. Jacobs}, \bibinfo{person}{Michael~I. Jordan}, \bibinfo{person}{Steven~J. Nowlan}, {and} \bibinfo{person}{Geoffrey~E. Hinton}.} \bibinfo{year}{1991}\natexlab{}.
\newblock \showarticletitle{Adaptive Mixtures of Local Experts}.
\newblock \bibinfo{journal}{\emph{Neural Computation}} \bibinfo{volume}{3}, \bibinfo{number}{1} (\bibinfo{year}{1991}).
\newblock


\bibitem[Kim et~al\mbox{.}(2016)]%
        {yong2016compression}
\bibfield{author}{\bibinfo{person}{Yong{-}Deok Kim}, \bibinfo{person}{Eunhyeok Park}, \bibinfo{person}{Sungjoo Yoo}, \bibinfo{person}{Taelim Choi}, \bibinfo{person}{Lu Yang}, {and} \bibinfo{person}{Dongjun Shin}.} \bibinfo{year}{2016}\natexlab{}.
\newblock \showarticletitle{Compression of Deep Convolutional Neural Networks for Fast and Low Power Mobile Applications}. In \bibinfo{booktitle}{\emph{International Conference on Learning Representations (ICLR)}}.
\newblock


\bibitem[Kirsch et~al\mbox{.}(2018)]%
        {kirsch2018modular}
\bibfield{author}{\bibinfo{person}{Louis Kirsch}, \bibinfo{person}{Julius Kunze}, {and} \bibinfo{person}{David Barber}.} \bibinfo{year}{2018}\natexlab{}.
\newblock \showarticletitle{Modular Networks: Learning to Decompose Neural Computation}. In \bibinfo{booktitle}{\emph{Proceedings of the 32nd International Conference on Neural Information Processing Systems (NeurIPS)}}.
\newblock


\bibitem[Krizhevsky et~al\mbox{.}(2009)]%
        {krizhevsky2009learning}
\bibfield{author}{\bibinfo{person}{Alex Krizhevsky}, \bibinfo{person}{Geoffrey Hinton}, {et~al\mbox{.}}} \bibinfo{year}{2009}\natexlab{}.
\newblock \showarticletitle{Learning multiple layers of features from tiny images}.
\newblock  (\bibinfo{year}{2009}).
\newblock


\bibitem[Laskaridis et~al\mbox{.}(2020)]%
        {laskaridis2020spinn}
\bibfield{author}{\bibinfo{person}{Stefanos Laskaridis}, \bibinfo{person}{Stylianos~I. Venieris}, \bibinfo{person}{Mario Almeida}, \bibinfo{person}{Ilias Leontiadis}, {and} \bibinfo{person}{Nicholas~D. Lane}.} \bibinfo{year}{2020}\natexlab{}.
\newblock \showarticletitle{SPINN: Synergistic Progressive Inference of Neural Networks over Device and Cloud}. In \bibinfo{booktitle}{\emph{Proceedings of the 26th Annual International Conference on Mobile Computing and Networking (MobiCom)}}.
\newblock


\bibitem[Li et~al\mbox{.}(2021)]%
        {li2021hermes}
\bibfield{author}{\bibinfo{person}{Ang Li}, \bibinfo{person}{Jingwei Sun}, \bibinfo{person}{Pengcheng Li}, \bibinfo{person}{Yu Pu}, \bibinfo{person}{Hai Li}, {and} \bibinfo{person}{Yiran Chen}.} \bibinfo{year}{2021}\natexlab{}.
\newblock \showarticletitle{Hermes: an efficient federated learning framework for heterogeneous mobile clients}. In \bibinfo{booktitle}{\emph{Proceedings of the 27th Annual International Conference on Mobile Computing and Networking (MobiCom)}}.
\newblock


\bibitem[Li and Wang(2019)]%
        {li2019fedmd}
\bibfield{author}{\bibinfo{person}{Daliang Li} {and} \bibinfo{person}{Junpu Wang}.} \bibinfo{year}{2019}\natexlab{}.
\newblock \showarticletitle{Fedmd: Heterogenous federated learning via model distillation}.
\newblock \bibinfo{journal}{\emph{arXiv preprint arXiv:1910.03581}} (\bibinfo{year}{2019}).
\newblock


\bibitem[Li et~al\mbox{.}(2020)]%
        {li2020federated}
\bibfield{author}{\bibinfo{person}{Tian Li}, \bibinfo{person}{Anit~Kumar Sahu}, \bibinfo{person}{Ameet Talwalkar}, {and} \bibinfo{person}{Virginia Smith}.} \bibinfo{year}{2020}\natexlab{}.
\newblock \showarticletitle{Federated learning: Challenges, methods, and future directions}.
\newblock \bibinfo{journal}{\emph{IEEE signal processing magazine}} \bibinfo{volume}{37}, \bibinfo{number}{3} (\bibinfo{year}{2020}).
\newblock


\bibitem[Li et~al\mbox{.}(2023)]%
        {li2023hierachical}
\bibfield{author}{\bibinfo{person}{Youpeng Li}, \bibinfo{person}{Xuyu Wang}, {and} \bibinfo{person}{Lingling An}.} \bibinfo{year}{2023}\natexlab{}.
\newblock \showarticletitle{Hierarchical Clustering-Based Personalized Federated Learning for Robust and Fair Human Activity Recognition}.
\newblock \bibinfo{journal}{\emph{Proc. ACM Interact. Mob. Wearable Ubiquitous Technol. (IMWUT)}} \bibinfo{volume}{7}, \bibinfo{number}{1} (\bibinfo{year}{2023}).
\newblock


\bibitem[Li and Hoiem(2017)]%
        {li2017learning}
\bibfield{author}{\bibinfo{person}{Zhizhong Li} {and} \bibinfo{person}{Derek Hoiem}.} \bibinfo{year}{2017}\natexlab{}.
\newblock \showarticletitle{Learning without forgetting}.
\newblock \bibinfo{journal}{\emph{IEEE transactions on pattern analysis and machine intelligence (TPAMI)}} \bibinfo{volume}{40}, \bibinfo{number}{12} (\bibinfo{year}{2017}).
\newblock


\bibitem[Lin et~al\mbox{.}(2017)]%
        {lin2017runtime}
\bibfield{author}{\bibinfo{person}{Ji Lin}, \bibinfo{person}{Yongming Rao}, \bibinfo{person}{Jiwen Lu}, {and} \bibinfo{person}{Jie Zhou}.} \bibinfo{year}{2017}\natexlab{}.
\newblock \showarticletitle{Runtime Neural Pruning}. In \bibinfo{booktitle}{\emph{Proceedings of the 31st International Conference on Neural Information Processing Systems (NeurIPS)}}.
\newblock


\bibitem[Lin et~al\mbox{.}(2020)]%
        {lin2020ensemble}
\bibfield{author}{\bibinfo{person}{Tao Lin}, \bibinfo{person}{Lingjing Kong}, \bibinfo{person}{Sebastian~U Stich}, {and} \bibinfo{person}{Martin Jaggi}.} \bibinfo{year}{2020}\natexlab{}.
\newblock \showarticletitle{Ensemble Distillation for Robust Model Fusion in Federated Learning}. In \bibinfo{booktitle}{\emph{Advances in Neural Information Processing Systems (NeurIPS)}}, Vol.~\bibinfo{volume}{33}.
\newblock


\bibitem[Liu et~al\mbox{.}(2022a)]%
        {liu2022distfl}
\bibfield{author}{\bibinfo{person}{Bingyan Liu}, \bibinfo{person}{Yifeng Cai}, \bibinfo{person}{Ziqi Zhang}, \bibinfo{person}{Yuanchun Li}, \bibinfo{person}{Leye Wang}, \bibinfo{person}{Ding Li}, \bibinfo{person}{Yao Guo}, {and} \bibinfo{person}{Xiangqun Chen}.} \bibinfo{year}{2022}\natexlab{a}.
\newblock \showarticletitle{DistFL: Distribution-Aware Federated Learning for Mobile Scenarios}.
\newblock \bibinfo{journal}{\emph{Proc. ACM Interact. Mob. Wearable Ubiquitous Technol. (IMWUT)}} \bibinfo{volume}{5}, \bibinfo{number}{4} (\bibinfo{year}{2022}).
\newblock


\bibitem[Liu et~al\mbox{.}(2019)]%
        {liu2019edge}
\bibfield{author}{\bibinfo{person}{Luyang Liu}, \bibinfo{person}{Hongyu Li}, {and} \bibinfo{person}{Marco Gruteser}.} \bibinfo{year}{2019}\natexlab{}.
\newblock \showarticletitle{Edge Assisted Real-Time Object Detection for Mobile Augmented Reality}. In \bibinfo{booktitle}{\emph{The 25th Annual International Conference on Mobile Computing and Networking (MobiCom)}}.
\newblock


\bibitem[Liu et~al\mbox{.}(2022b)]%
        {liu2022no}
\bibfield{author}{\bibinfo{person}{Ruixuan Liu}, \bibinfo{person}{Fangzhao Wu}, \bibinfo{person}{Chuhan Wu}, \bibinfo{person}{Yanlin Wang}, \bibinfo{person}{Lingjuan Lyu}, \bibinfo{person}{Hong Chen}, {and} \bibinfo{person}{Xing Xie}.} \bibinfo{year}{2022}\natexlab{b}.
\newblock \showarticletitle{No One Left Behind: Inclusive Federated Learning over Heterogeneous Devices}. In \bibinfo{booktitle}{\emph{Proceedings of the 28th ACM SIGKDD Conference on Knowledge Discovery and Data Mining (SIGKDD)}}.
\newblock


\bibitem[Liu et~al\mbox{.}(2021)]%
        {liu2021adaspring}
\bibfield{author}{\bibinfo{person}{Sicong Liu}, \bibinfo{person}{Bin Guo}, \bibinfo{person}{Ke Ma}, \bibinfo{person}{Zhiwen Yu}, {and} \bibinfo{person}{Junzhao Du}.} \bibinfo{year}{2021}\natexlab{}.
\newblock \showarticletitle{AdaSpring: Context-Adaptive and Runtime-Evolutionary Deep Model Compression for Mobile Applications}.
\newblock \bibinfo{journal}{\emph{Proc. ACM Interact. Mob. Wearable Ubiquitous Technol. (IMWUT)}} \bibinfo{volume}{5}, \bibinfo{number}{1} (\bibinfo{year}{2021}).
\newblock


\bibitem[Lv et~al\mbox{.}(2022)]%
        {chengfei2022walle}
\bibfield{author}{\bibinfo{person}{Chengfei Lv}, \bibinfo{person}{Chaoyue Niu}, \bibinfo{person}{Renjie Gu}, \bibinfo{person}{Xiaotang Jiang}, \bibinfo{person}{Zhaode Wang}, \bibinfo{person}{Bin Liu}, \bibinfo{person}{Ziqi Wu}, \bibinfo{person}{Qiulin Yao}, \bibinfo{person}{Congyu Huang}, \bibinfo{person}{Panos Huang}, \bibinfo{person}{Tao Huang}, \bibinfo{person}{Hui Shu}, \bibinfo{person}{Jinde Song}, \bibinfo{person}{Bin Zou}, \bibinfo{person}{Peng Lan}, \bibinfo{person}{Guohuan Xu}, \bibinfo{person}{Fei Wu}, \bibinfo{person}{Shaojie Tang}, \bibinfo{person}{Fan Wu}, {and} \bibinfo{person}{Guihai Chen}.} \bibinfo{year}{2022}\natexlab{}.
\newblock \showarticletitle{Walle: An {End-to-End}, {General-Purpose}, and {Large-Scale} Production System for {Device-Cloud} Collaborative Machine Learning}. In \bibinfo{booktitle}{\emph{16th USENIX Symposium on Operating Systems Design and Implementation (OSDI 22)}}.
\newblock


\bibitem[Ma et~al\mbox{.}(2018b)]%
        {modeling2018ma}
\bibfield{author}{\bibinfo{person}{Jiaqi Ma}, \bibinfo{person}{Zhe Zhao}, \bibinfo{person}{Xinyang Yi}, \bibinfo{person}{Jilin Chen}, \bibinfo{person}{Lichan Hong}, {and} \bibinfo{person}{Ed~H. Chi}.} \bibinfo{year}{2018}\natexlab{b}.
\newblock \showarticletitle{Modeling Task Relationships in Multi-Task Learning with Multi-Gate Mixture-of-Experts}. In \bibinfo{booktitle}{\emph{Proceedings of the 24th ACM SIGKDD International Conference on Knowledge Discovery \& Data Mining (SIGKDD)}}.
\newblock


\bibitem[Ma et~al\mbox{.}(2018a)]%
        {ma2018shufflenet}
\bibfield{author}{\bibinfo{person}{Ningning Ma}, \bibinfo{person}{Xiangyu Zhang}, \bibinfo{person}{Hai-Tao Zheng}, {and} \bibinfo{person}{Jian Sun}.} \bibinfo{year}{2018}\natexlab{a}.
\newblock \showarticletitle{Shufflenet v2: Practical guidelines for efficient cnn architecture design}. In \bibinfo{booktitle}{\emph{Proceedings of the European conference on computer vision (ECCV)}}.
\newblock


\bibitem[Masoudnia and Ebrahimpour(2014)]%
        {masoudnia2014mixture}
\bibfield{author}{\bibinfo{person}{Saeed Masoudnia} {and} \bibinfo{person}{Reza Ebrahimpour}.} \bibinfo{year}{2014}\natexlab{}.
\newblock \showarticletitle{Mixture of experts: a literature survey}.
\newblock \bibinfo{journal}{\emph{The Artificial Intelligence Review}} \bibinfo{volume}{42}, \bibinfo{number}{2} (\bibinfo{year}{2014}).
\newblock


\bibitem[McMahan et~al\mbox{.}(2017)]%
        {mcmahan2017communication}
\bibfield{author}{\bibinfo{person}{Brendan McMahan}, \bibinfo{person}{Eider Moore}, \bibinfo{person}{Daniel Ramage}, \bibinfo{person}{Seth Hampson}, {and} \bibinfo{person}{Blaise~Aguera y Arcas}.} \bibinfo{year}{2017}\natexlab{}.
\newblock \showarticletitle{Communication-efficient learning of deep networks from decentralized data}. In \bibinfo{booktitle}{\emph{Artificial intelligence and statistics (AISTATS)}}. PMLR.
\newblock


\bibitem[Misra et~al\mbox{.}(2016)]%
        {crossstitch2016misra}
\bibfield{author}{\bibinfo{person}{Ishan Misra}, \bibinfo{person}{Abhinav Shrivastava}, \bibinfo{person}{Abhinav Gupta}, {and} \bibinfo{person}{Martial Hebert}.} \bibinfo{year}{2016}\natexlab{}.
\newblock \showarticletitle{Cross-Stitch Networks for Multi-task Learning}. In \bibinfo{booktitle}{\emph{2016 IEEE Conference on Computer Vision and Pattern Recognition (CVPR)}}.
\newblock


\bibitem[Niu et~al\mbox{.}(2020)]%
        {niu2020billion}
\bibfield{author}{\bibinfo{person}{Chaoyue Niu}, \bibinfo{person}{Fan Wu}, \bibinfo{person}{Shaojie Tang}, \bibinfo{person}{Lifeng Hua}, \bibinfo{person}{Rongfei Jia}, \bibinfo{person}{Chengfei Lv}, \bibinfo{person}{Zhihua Wu}, {and} \bibinfo{person}{Guihai Chen}.} \bibinfo{year}{2020}\natexlab{}.
\newblock \showarticletitle{Billion-Scale Federated Learning on Mobile Clients: A Submodel Design with Tunable Privacy}. In \bibinfo{booktitle}{\emph{Proceedings of the 26th Annual International Conference on Mobile Computing and Networking (MobiCom)}}.
\newblock


\bibitem[Ouyang et~al\mbox{.}(2021)]%
        {Ouyang2021clusterfl}
\bibfield{author}{\bibinfo{person}{Xiaomin Ouyang}, \bibinfo{person}{Zhiyuan Xie}, \bibinfo{person}{Jiayu Zhou}, \bibinfo{person}{Jianwei Huang}, {and} \bibinfo{person}{Guoliang Xing}.} \bibinfo{year}{2021}\natexlab{}.
\newblock \showarticletitle{ClusterFL: A Similarity-Aware Federated Learning System for Human Activity Recognition}. In \bibinfo{booktitle}{\emph{Proceedings of the 19th Annual International Conference on Mobile Systems, Applications, and Services (MobiSys)}}.
\newblock


\bibitem[Padmanabhan et~al\mbox{.}(2023)]%
        {Padmanabhan2022gemel}
\bibfield{author}{\bibinfo{person}{Arthi Padmanabhan}, \bibinfo{person}{Neil Agarwal}, \bibinfo{person}{Anand Iyer}, \bibinfo{person}{Ganesh Ananthanarayanan}, \bibinfo{person}{Yuanchao Shu}, \bibinfo{person}{Nikolaos Karianakis}, \bibinfo{person}{Guoqing~Harry Xu}, {and} \bibinfo{person}{Ravi Netravali}.} \bibinfo{year}{2023}\natexlab{}.
\newblock \showarticletitle{GEMEL: Model Merging for Memory-Efficient, Real-Time Video Analytics at the Edge}. In \bibinfo{booktitle}{\emph{Symposium on Networked Systems Design and Implementation (NSDI)}}.
\newblock


\bibitem[Paszke et~al\mbox{.}(2019)]%
        {paszke2019pytorch}
\bibfield{author}{\bibinfo{person}{Adam Paszke}, \bibinfo{person}{Sam Gross}, \bibinfo{person}{Francisco Massa}, \bibinfo{person}{Adam Lerer}, \bibinfo{person}{James Bradbury}, \bibinfo{person}{Gregory Chanan}, \bibinfo{person}{Trevor Killeen}, \bibinfo{person}{Zeming Lin}, \bibinfo{person}{Natalia Gimelshein}, \bibinfo{person}{Luca Antiga}, \bibinfo{person}{Alban Desmaison}, \bibinfo{person}{Andreas K\"{o}pf}, \bibinfo{person}{Edward Yang}, \bibinfo{person}{Zach DeVito}, \bibinfo{person}{Martin Raison}, \bibinfo{person}{Alykhan Tejani}, \bibinfo{person}{Sasank Chilamkurthy}, \bibinfo{person}{Benoit Steiner}, \bibinfo{person}{Lu Fang}, \bibinfo{person}{Junjie Bai}, {and} \bibinfo{person}{Soumith Chintala}.} \bibinfo{year}{2019}\natexlab{}.
\newblock \showarticletitle{PyTorch: An Imperative Style, High-Performance Deep Learning Library}. In \bibinfo{booktitle}{\emph{Proceedings of the 33rd International Conference on Neural Information Processing Systems (NeurIPS)}}.
\newblock


\bibitem[Ravi and Kozareva(2018)]%
        {ravi2018self}
\bibfield{author}{\bibinfo{person}{Sujith Ravi} {and} \bibinfo{person}{Zornitsa Kozareva}.} \bibinfo{year}{2018}\natexlab{}.
\newblock \showarticletitle{Self-Governing Neural Networks for On-Device Short Text Classification}. In \bibinfo{booktitle}{\emph{Proceedings of the 2018 Conference on Empirical Methods in Natural Language Processing (EMNLP)}}.
\newblock


\bibitem[Sandler et~al\mbox{.}(2018)]%
        {sandler2018mobilenetv2}
\bibfield{author}{\bibinfo{person}{Mark Sandler}, \bibinfo{person}{Andrew Howard}, \bibinfo{person}{Menglong Zhu}, \bibinfo{person}{Andrey Zhmoginov}, {and} \bibinfo{person}{Liang-Chieh Chen}.} \bibinfo{year}{2018}\natexlab{}.
\newblock \showarticletitle{Mobilenetv2: Inverted residuals and linear bottlenecks}. In \bibinfo{booktitle}{\emph{Proceedings of the IEEE conference on computer vision and pattern recognition (CVPR)}}.
\newblock


\bibitem[Shazeer et~al\mbox{.}(2017)]%
        {shazeer2017outrageously}
\bibfield{author}{\bibinfo{person}{Noam Shazeer}, \bibinfo{person}{Azalia Mirhoseini}, \bibinfo{person}{Krzysztof Maziarz}, \bibinfo{person}{Andy Davis}, \bibinfo{person}{Quoc Le}, \bibinfo{person}{Geoffrey Hinton}, {and} \bibinfo{person}{Jeff Dean}.} \bibinfo{year}{2017}\natexlab{}.
\newblock \showarticletitle{Outrageously Large Neural Networks: The Sparsely-Gated Mixture-of-Experts Layer}. In \bibinfo{booktitle}{\emph{International Conference on Learning Representations (ICLR)}}.
\newblock


\bibitem[Simonyan and Zisserman(2015)]%
        {simonyan2014very}
\bibfield{author}{\bibinfo{person}{Karen Simonyan} {and} \bibinfo{person}{Andrew Zisserman}.} \bibinfo{year}{2015}\natexlab{}.
\newblock \showarticletitle{Very deep convolutional networks for large-scale image recognition}. In \bibinfo{booktitle}{\emph{International Conference on Learning Representations (ICLR)}}.
\newblock


\bibitem[Tan and Le(2021)]%
        {tan2021efficientnetv2}
\bibfield{author}{\bibinfo{person}{Mingxing Tan} {and} \bibinfo{person}{Quoc Le}.} \bibinfo{year}{2021}\natexlab{}.
\newblock \showarticletitle{Efficientnetv2: Smaller models and faster training}. In \bibinfo{booktitle}{\emph{International conference on machine learning (ICML)}}.
\newblock


\bibitem[Teerapittayanon et~al\mbox{.}(2016)]%
        {teerapittayanon2016branchynet}
\bibfield{author}{\bibinfo{person}{Surat Teerapittayanon}, \bibinfo{person}{Bradley McDanel}, {and} \bibinfo{person}{Hsiang-Tsung Kung}.} \bibinfo{year}{2016}\natexlab{}.
\newblock \showarticletitle{Branchynet: Fast inference via early exiting from deep neural networks}. In \bibinfo{booktitle}{\emph{2016 23rd International Conference on Pattern Recognition (ICPR)}}. IEEE.
\newblock


\bibitem[Tu et~al\mbox{.}(2021)]%
        {tu2021feddl}
\bibfield{author}{\bibinfo{person}{Linlin Tu}, \bibinfo{person}{Xiaomin Ouyang}, \bibinfo{person}{Jiayu Zhou}, \bibinfo{person}{Yuze He}, {and} \bibinfo{person}{Guoliang Xing}.} \bibinfo{year}{2021}\natexlab{}.
\newblock \showarticletitle{FedDL: Federated Learning via Dynamic Layer Sharing for Human Activity Recognition}. In \bibinfo{booktitle}{\emph{Proceedings of the 19th ACM Conference on Embedded Networked Sensor Systems (SenSys)}}.
\newblock


\bibitem[Vaswani et~al\mbox{.}(2017)]%
        {vaswani2017attention}
\bibfield{author}{\bibinfo{person}{Ashish Vaswani}, \bibinfo{person}{Noam Shazeer}, \bibinfo{person}{Niki Parmar}, \bibinfo{person}{Jakob Uszkoreit}, \bibinfo{person}{Llion Jones}, \bibinfo{person}{Aidan~N Gomez}, \bibinfo{person}{\L~ukasz Kaiser}, {and} \bibinfo{person}{Illia Polosukhin}.} \bibinfo{year}{2017}\natexlab{}.
\newblock \showarticletitle{Attention is All you Need}. In \bibinfo{booktitle}{\emph{Advances in Neural Information Processing Systems (NeurIPS)}}.
\newblock


\bibitem[Vepakomma et~al\mbox{.}(2018)]%
        {Vepakomma2018SplitLF}
\bibfield{author}{\bibinfo{person}{Praneeth Vepakomma}, \bibinfo{person}{Otkrist Gupta}, \bibinfo{person}{Tristan Swedish}, {and} \bibinfo{person}{Ramesh Raskar}.} \bibinfo{year}{2018}\natexlab{}.
\newblock \showarticletitle{Split learning for health: Distributed deep learning without sharing raw patient data}.
\newblock \bibinfo{journal}{\emph{ArXiv}}  \bibinfo{volume}{abs/1812.00564} (\bibinfo{year}{2018}).
\newblock


\bibitem[Virtanen et~al\mbox{.}(2020)]%
        {2020SciPy-NMeth}
\bibfield{author}{\bibinfo{person}{Pauli Virtanen}, \bibinfo{person}{Ralf Gommers}, \bibinfo{person}{Travis~E. Oliphant}, \bibinfo{person}{Matt Haberland}, \bibinfo{person}{Tyler Reddy}, \bibinfo{person}{David Cournapeau}, \bibinfo{person}{Evgeni Burovski}, \bibinfo{person}{Pearu Peterson}, \bibinfo{person}{Warren Weckesser}, \bibinfo{person}{Jonathan Bright}, \bibinfo{person}{St{\'e}fan~J. {van der Walt}}, \bibinfo{person}{Matthew Brett}, \bibinfo{person}{Joshua Wilson}, \bibinfo{person}{K.~Jarrod Millman}, \bibinfo{person}{Nikolay Mayorov}, \bibinfo{person}{Andrew R.~J. Nelson}, \bibinfo{person}{Eric Jones}, \bibinfo{person}{Robert Kern}, \bibinfo{person}{Eric Larson}, \bibinfo{person}{C~J Carey}, \bibinfo{person}{{\.I}lhan Polat}, \bibinfo{person}{Yu Feng}, \bibinfo{person}{Eric~W. Moore}, \bibinfo{person}{Jake {VanderPlas}}, \bibinfo{person}{Denis Laxalde}, \bibinfo{person}{Josef Perktold}, \bibinfo{person}{Robert Cimrman}, \bibinfo{person}{Ian Henriksen}, \bibinfo{person}{E.~A. Quintero}, \bibinfo{person}{Charles~R. Harris}, \bibinfo{person}{Anne~M. Archibald}, \bibinfo{person}{Ant{\^o}nio~H. Ribeiro}, \bibinfo{person}{Fabian Pedregosa}, \bibinfo{person}{Paul {van Mulbregt}}, {and} \bibinfo{person}{{SciPy 1.0 Contributors}}.} \bibinfo{year}{2020}\natexlab{}.
\newblock \showarticletitle{{{SciPy} 1.0: Fundamental Algorithms for Scientific Computing in Python}}.
\newblock \bibinfo{journal}{\emph{Nature Methods}}  \bibinfo{volume}{17} (\bibinfo{year}{2020}).
\newblock
\urldef\tempurl%
\url{https://doi.org/10.1038/s41592-019-0686-2}
\showDOI{\tempurl}


\bibitem[Wang et~al\mbox{.}(2021)]%
        {wang2021context}
\bibfield{author}{\bibinfo{person}{Hongli Wang}, \bibinfo{person}{Bin Guo}, \bibinfo{person}{Jiaqi Liu}, \bibinfo{person}{Sicong Liu}, \bibinfo{person}{Yungang Wu}, {and} \bibinfo{person}{Zhiwen Yu}.} \bibinfo{year}{2021}\natexlab{}.
\newblock \showarticletitle{Context-Aware Adaptive Surgery: A Fast and Effective Framework for Adaptative Model Partition}.
\newblock \bibinfo{journal}{\emph{Proc. ACM Interact. Mob. Wearable Ubiquitous Technol. (IMWUT)}} \bibinfo{volume}{5}, \bibinfo{number}{3} (\bibinfo{year}{2021}).
\newblock


\bibitem[Wang and Joshi(2019)]%
        {wang2019adaptive}
\bibfield{author}{\bibinfo{person}{Jianyu Wang} {and} \bibinfo{person}{Gauri Joshi}.} \bibinfo{year}{2019}\natexlab{}.
\newblock \showarticletitle{Adaptive communication strategies to achieve the best error-runtime trade-off in local-update SGD}.
\newblock \bibinfo{journal}{\emph{Proceedings of Machine Learning and Systems (MLSys)}}  \bibinfo{volume}{1} (\bibinfo{year}{2019}).
\newblock


\bibitem[Wang et~al\mbox{.}(2018)]%
        {wang2018skipnet}
\bibfield{author}{\bibinfo{person}{Xin Wang}, \bibinfo{person}{Fisher Yu}, \bibinfo{person}{Zi-Yi Dou}, \bibinfo{person}{Trevor Darrell}, {and} \bibinfo{person}{Joseph~E Gonzalez}.} \bibinfo{year}{2018}\natexlab{}.
\newblock \showarticletitle{Skipnet: Learning dynamic routing in convolutional networks}. In \bibinfo{booktitle}{\emph{Proceedings of the European Conference on Computer Vision (ECCV)}}.
\newblock


\bibitem[Warden(2018)]%
        {warden2018speech}
\bibfield{author}{\bibinfo{person}{Pete Warden}.} \bibinfo{year}{2018}\natexlab{}.
\newblock \showarticletitle{Speech commands: A dataset for limited-vocabulary speech recognition}.
\newblock \bibinfo{journal}{\emph{arXiv preprint arXiv:1804.03209}} (\bibinfo{year}{2018}).
\newblock


\bibitem[Wen et~al\mbox{.}(2023)]%
        {wen2023adaptivenet}
\bibfield{author}{\bibinfo{person}{Hao Wen}, \bibinfo{person}{Yuanchun Li}, \bibinfo{person}{Zunshuai Zhang}, \bibinfo{person}{Shiqi Jiang}, \bibinfo{person}{Xiaozhou Ye}, \bibinfo{person}{Ye Ouyang}, \bibinfo{person}{Ya-Qin Zhang}, {and} \bibinfo{person}{Yunxin Liu}.} \bibinfo{year}{2023}\natexlab{}.
\newblock \showarticletitle{AdaptiveNet: Post-deployment Neural Architecture Adaptation for Diverse Edge Environments}. In \bibinfo{booktitle}{\emph{Proceedings of the 29th Annual International Conference on Mobile Computing and Networking (MobiCom)}}.
\newblock


\bibitem[Wen et~al\mbox{.}(2016)]%
        {wei2016learning}
\bibfield{author}{\bibinfo{person}{Wei Wen}, \bibinfo{person}{Chunpeng Wu}, \bibinfo{person}{Yandan Wang}, \bibinfo{person}{Yiran Chen}, {and} \bibinfo{person}{Hai Li}.} \bibinfo{year}{2016}\natexlab{}.
\newblock \showarticletitle{Learning Structured Sparsity in Deep Neural Networks}. In \bibinfo{booktitle}{\emph{Proceedings of the 30th International Conference on Neural Information Processing Systems (NeurIPS)}}.
\newblock


\bibitem[Wu et~al\mbox{.}(2018)]%
        {wu2018blockdrop}
\bibfield{author}{\bibinfo{person}{Zuxuan Wu}, \bibinfo{person}{Tushar Nagarajan}, \bibinfo{person}{Abhishek Kumar}, \bibinfo{person}{Steven Rennie}, \bibinfo{person}{Larry~S Davis}, \bibinfo{person}{Kristen Grauman}, {and} \bibinfo{person}{Rogerio Feris}.} \bibinfo{year}{2018}\natexlab{}.
\newblock \showarticletitle{Blockdrop: Dynamic inference paths in residual networks}. In \bibinfo{booktitle}{\emph{Proceedings of the IEEE conference on computer vision and pattern recognition (CVPR)}}.
\newblock


\bibitem[Yao et~al\mbox{.}(2021)]%
        {yao2021device}
\bibfield{author}{\bibinfo{person}{Jiangchao Yao}, \bibinfo{person}{Feng Wang}, \bibinfo{person}{Kunyang Jia}, \bibinfo{person}{Bo Han}, \bibinfo{person}{Jingren Zhou}, {and} \bibinfo{person}{Hongxia Yang}.} \bibinfo{year}{2021}\natexlab{}.
\newblock \showarticletitle{Device-Cloud Collaborative Learning for Recommendation}. In \bibinfo{booktitle}{\emph{Proceedings of the 27th ACM SIGKDD Conference on Knowledge Discovery \& Data Mining (SIGKDD)}}.
\newblock


\bibitem[Yao et~al\mbox{.}(2022)]%
        {yao2022edge}
\bibfield{author}{\bibinfo{person}{Jiangchao Yao}, \bibinfo{person}{Shengyu Zhang}, \bibinfo{person}{Yang Yao}, \bibinfo{person}{Feng Wang}, \bibinfo{person}{Jianxin Ma}, \bibinfo{person}{Jianwei Zhang}, \bibinfo{person}{Yunfei Chu}, \bibinfo{person}{Luo Ji}, \bibinfo{person}{Kunyang Jia}, \bibinfo{person}{Tao Shen}, \bibinfo{person}{Anpeng Wu}, \bibinfo{person}{Fengda Zhang}, \bibinfo{person}{Ziqi Tan}, \bibinfo{person}{Kun Kuang}, \bibinfo{person}{Chao Wu}, \bibinfo{person}{Fei Wu}, \bibinfo{person}{Jingren Zhou}, {and} \bibinfo{person}{Hongxia Yang}.} \bibinfo{year}{2022}\natexlab{}.
\newblock \showarticletitle{Edge-Cloud Polarization and Collaboration: A Comprehensive Survey for AI}.
\newblock \bibinfo{journal}{\emph{IEEE Transactions on Knowledge and Data Engineering (TKDE)}} (\bibinfo{year}{2022}).
\newblock


\bibitem[Zeng et~al\mbox{.}(2017)]%
        {zeng2017mobiledeeppill}
\bibfield{author}{\bibinfo{person}{Xiao Zeng}, \bibinfo{person}{Kai Cao}, {and} \bibinfo{person}{Mi Zhang}.} \bibinfo{year}{2017}\natexlab{}.
\newblock \showarticletitle{MobileDeepPill: A Small-Footprint Mobile Deep Learning System for Recognizing Unconstrained Pill Images}. In \bibinfo{booktitle}{\emph{Proceedings of the 15th Annual International Conference on Mobile Systems, Applications, and Services (MobiSys)}}.
\newblock


\bibitem[Zhang et~al\mbox{.}(2018)]%
        {zhang2018deep}
\bibfield{author}{\bibinfo{person}{Lei Zhang}, \bibinfo{person}{Shuai Wang}, {and} \bibinfo{person}{Bing Liu}.} \bibinfo{year}{2018}\natexlab{}.
\newblock \showarticletitle{Deep learning for sentiment analysis: {A} survey}.
\newblock \bibinfo{journal}{\emph{Wiley Interdisciplinary Reviews: Data Mining and Knowledge Discovery}} \bibinfo{volume}{8}, \bibinfo{number}{4} (\bibinfo{year}{2018}).
\newblock


\bibitem[Zhao et~al\mbox{.}(2018)]%
        {Zhao2018FederatedLW}
\bibfield{author}{\bibinfo{person}{Yue Zhao}, \bibinfo{person}{Meng Li}, \bibinfo{person}{Liangzhen Lai}, \bibinfo{person}{Naveen Suda}, \bibinfo{person}{Damon Civin}, {and} \bibinfo{person}{Vikas Chandra}.} \bibinfo{year}{2018}\natexlab{}.
\newblock \showarticletitle{Federated Learning with Non-IID Data}.
\newblock \bibinfo{journal}{\emph{arXiv preprint arXiv:1806.00582}} (\bibinfo{year}{2018}).
\newblock


\end{thebibliography}

\end{document}